\documentclass{article}

\usepackage{microtype}
\usepackage{graphicx}
\usepackage{subfigure}
\usepackage{booktabs} 
    
\usepackage[hidelinks]{hyperref}

\usepackage{float}
\usepackage{multirow}

\usepackage{amsmath}
\usepackage{amssymb}
\usepackage{mathtools}
\usepackage{amsthm}
\usepackage{tabularx}
\usepackage{tikz}
\usetikzlibrary{shapes.geometric, arrows}
\usepackage{graphicx}
\usepackage{subcaption}
\usepackage{pifont}
\usepackage{array}
\usepackage{xcolor}
\usepackage{listings}
\usepackage{algorithm}
\usepackage[noend]{algpseudocode}
\usepackage{svg}
\usepackage{bbm}
\usepackage{enumitem}

\usepackage[capitalize,noabbrev]{cleveref}

\usepackage{wrapfig}
\usepackage{placeins}

\theoremstyle{plain}

\theoremstyle{definition}

\theoremstyle{remark}

\usepackage[disable,textsize=tiny]{todonotes}
\usepackage{thmtools} 
\usepackage{thm-restate}

\usepackage[numbers, compress]{natbib}

\usepackage[preprint]{neurips_2026}

\usepackage[utf8]{inputenc} 
\usepackage[T1]{fontenc}    
\usepackage{hyperref}       
\usepackage{url}            
\usepackage{booktabs}       
\usepackage{amsfonts}       
\usepackage{nicefrac}       
\usepackage{microtype}      
\usepackage{xcolor}         

\definecolor{promptbg}{RGB}{248,248,248}
\definecolor{promptframe}{RGB}{210,210,210}

\lstdefinestyle{promptstyle}{
    basicstyle=\ttfamily\footnotesize,
    backgroundcolor=\color{promptbg},
    frame=single,
    rulecolor=\color{promptframe},
    framerule=0.4pt,
    xleftmargin=0.5em,
    xrightmargin=0.5em,
    framexleftmargin=0.5em,
    framexrightmargin=0.5em,
    breaklines=true,
    breakatwhitespace=true,
    columns=fullflexible,
    keepspaces=true,
    showstringspaces=false,
    aboveskip=0.6em,
    belowskip=0.4em
}

\newcommand{\utkarsh}[1]{}
\newcommand{\pengfei}[1]{}

\title{Reinforcement Learning with Verifiable Physics: Post-training LLMs with Continuous Rewards}

\author{
  Pengfei Cai
  \thanks{Equal contribution. Order decided by coin toss.} \\
  Massachusetts Institute of Technology
  \And
  Utkarsh Utkarsh\footnotemark[1] \\
  Massachusetts Institute of Technology
  \And
  Alan Edelman \\
  Massachusetts Institute of Technology
  \And
  Christopher Vincent Rackauckas
  \thanks{Corresponding authors:
  \texttt{crackauc@mit.edu}, \texttt{rafagb@mit.edu}} \\
  Massachusetts Institute of Technology
  \And
  Rafael Gomez-Bombarelli\footnotemark[2] \\
  Massachusetts Institute of Technology
}

\begin{document}
\raggedbottom

\maketitle

\begin{abstract}
Partial differential equations (PDEs) are foundational to modeling in science and engineering, but constructing reliable numerical solvers remains labor-intensive, demanding expert knowledge of discretization schemes, stability conditions, and boundary treatments. Recent work has begun to frame PDE solving as a code-generation task for large language models (LLMs), yet existing approaches operate primarily at inference time: relying on prompting, debugging, self-refinement, and test-time scaling rather than adapting the model itself. In parallel, reinforcement learning with verifiable rewards has emerged as a powerful post-training paradigm for code and math reasoning, but its verifiers are typically binary: a compiler runs, or a test passes. Such signals discard the graded structure of scientific correctness, where two solvers may both execute and yet differ in solution accuracy by orders of magnitude. In this work, we introduce RLVP: Reinforcement Learning with Verifiable Physics, an RL post-training framework for multi-PDE solver code generation. RLVP addresses this verifiability gap with a hybrid verifier: hard program-validity checks ensure executability, while continuous physics rewards score function-space accuracy and
PDE-residual consistency. A single policy is post-trained across diverse PDE families spanning hyperbolic, parabolic, elliptic, and incompressible-flow systems. RLVP improves over both pre-trained and supervised-only baselines on PDE benchmarks, and shows zero-shot improvement transfer to held-out PDEs. We show that a smaller LLM post-trained with RLVP can outperform prompting a frontier model on in-distribution PDE solver generation. The trained policy shows evidence of compositionality in numerical motifs: it recombines stencils, time-stepping schemes, and boundary-handling primitives learned from the PDEs used in training into generated solvers for unseen PDE problems.
\end{abstract}

\section{Introduction}

Partial differential equations (PDEs) are a central language of the physical sciences,
governing systems in fluid dynamics, heat transfer, electromagnetism, structural
mechanics, and beyond \citep{evans2022partial, leveque2002finite,
quarteroni1994numerical}. Decades of scientific computing have produced powerful solver
families, including finite-difference, finite-volume, finite-element, spectral, implicit,
and operator-splitting schemes \citep{leveque2002finite, quarteroni1994numerical,
trefethen2000spectral, ascher1998computer}. Yet writing a reliable solver remains a
specialized task: correctness depends on discretization, stability constraints, boundary
treatment, time integration, solver tolerances, and physical invariants. Code can compile,
return arrays of the right shape, and still violate a CFL condition, mishandle a boundary,
introduce excessive diffusion, or produce a qualitatively wrong solution.

Large language models (LLMs) have become strong code generators and coding agents
\citep{chen2021evaluating, hendrycks2021measuring, austin2021program, le2022coderl,
yao2022react, gehring2024rlef}, raising the question of whether PDE solver construction
can be automated as code generation. Recent work suggests that this direction is
promising but incomplete: LLMs can produce plausible scientific code, but reliable solver
accuracy often requires domain-informed prompting, execution feedback, debugging loops,
self-refinement, agentic orchestration, or test-time scaling
\citep{li2025codepde, gaonkar2025sciml, du2026autonumerics, deotale2026all,
liu2025pde, wuwu2025pinnsagent, snell2024scaling, wu2024inference}. These inference-time
methods can improve individual outputs, but they do not amortize numerical reliability
into the model's solver-writing distribution. We instead ask whether the experience of
executing and grading PDE solvers can be converted into model parameter updates.

Reinforcement learning with verifiable rewards (RLVR) provides a natural starting point:
models can improve when trained against rule-based execution feedback rather than
imitation alone \citep{shao2024deepseekmath, guo2025deepseek, le2022coderl,
uesato2022solving}. However, standard RLVR verifiers are usually binary: an answer
matches, a unit test passes, or a format constraint is satisfied. PDE solver generation is
graded rather than binary. A generated program may be executable and finite while still
being unstable, overly diffusive, dispersive, biased, or inconsistent with the governing
equation; two valid programs may differ by orders of magnitude in solution error. This
graded structure makes PDE simulation an unusually rich verifiable environment: predicted
trajectories can be compared to hidden numerical references, checked for physics residual
consistency, and evaluated across held-out grids, parameters, initial and boundary
conditions.

We introduce \emph{Reinforcement Learning with Verifiable Physics} (RLVP), a
post-training framework for PDE solver code generation. Starting from supervised
fine-tuning on a curated bank of teacher solvers, we apply GRPO
\citep{shao2024deepseekmath} with a hybrid verifier: hard execution-validity gates ensure
program feasibility, while continuous function-space rewards score generated solutions
against hidden references with residual-consistency regularization. This moves physical
feedback from post-hoc evaluation into the training loop. A single policy is trained
jointly across eight PDE families spanning hyperbolic transport, parabolic
reaction--diffusion, steady elliptic problems, and incompressible Navier--Stokes.

Our contributions are:
\begin{enumerate}\itemsep0pt
    \item \textbf{Hybrid binary--continuous verifier for PDE solver code.}
    We design an RLVR-style verifier whose complete reward combines hard program validity \(V\), a continuous physical accuracy reward \(R_{\mathrm{traj}}\) based on function-space trajectory error, and a physics residual
    consistency reward \(R_{\mathrm{phys}}\).
    The verifier is computed deterministically from numerical reference solutions and PDE residuals, with no learned reward model in the loop. The hybrid reward outperforms a binary-validity-only baseline by 8--13 percentage points while reducing nRMSE by $38\%$, demonstrating that continuous physical accuracy contributes substantially beyond binary execution feedback.
    
    \item \textbf{A multi-PDE solver-bank and open post-training recipe.}
    We curate a solver-code dataset spanning 8 PDE families and 71 standard numerical
    schemes, including finite-difference stencils, conservative updates, CFL substepping,
    sparse linear solves, projection steps, and boundary conditions handling. We pair this
    dataset with a reproducible RLVP post-training pipeline built on
    vLLM~\citep{kwon2023efficient} and verl~\citep{sheng2025hybridflow}, which we will
    release together with the verifier environment and training recipes. Unlike prompt
    engineering approaches that require detailed PDE-specific domain knowledge, our
    pipeline uses simpler prompts, making post-trained models closer to useful scientific
    computing agents for broader audience.

    \item \textbf{Multi-PDE post-training.}
    A single policy is trained jointly across PDE families spanning hyperbolic, parabolic,
    elliptic, and incompressible-flow systems in 1D and 2D. To our knowledge, this is the
    first verifiable-reward RL pipeline that post-trains one model across PDE families
    rather than specializing to a single equation or narrow benchmark distribution.
    
    \item \textbf{Small RLVP-trained models can outperform larger static prompting baselines.}
    Across model sizes (3B, 7B, 14B), the post-trained policy matches or exceeds
    substantially larger frontier baselines under the same direct-generation protocol.
    
    \item \textbf{Selective zero-shot cross-PDE transfer.}
    The trained policy improves solver code generation on held-out PDEs absent from training, especially by recombining or extending numerical structures, indicating that RLVP post-training can instill transferable numerical-programming motifs rather than per-prompt template memorization.

\end{enumerate}

\begin{figure}[H]
    \centering
    \includegraphics[width=0.85\linewidth]{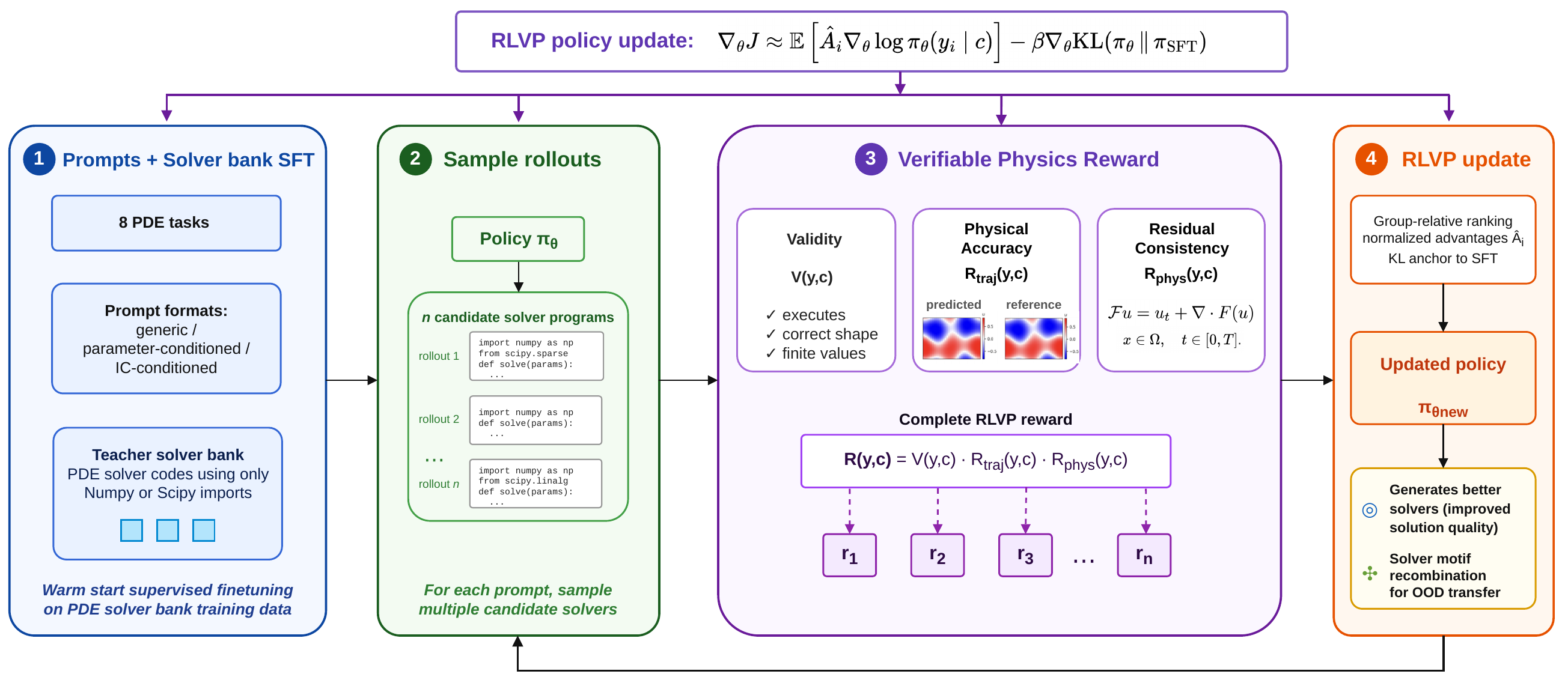}
    \caption{\textbf{Overview of RLVP for multi-PDE solver code generation.} We warm-start the policy with supervised fine-tuning (SFT) on a multi-PDE solver bank across diverse PDE problems. During RL, sampled solver programs are executed by the RLVP verifier and scored by a reward combining validity, physical accuracy against hidden numerical references, and physics residual consistency. Group-relative advantages update the policy with a KL anchor to the SFT model.}
    \label{fig:figure1}
\end{figure}

\vspace{-10px}
\section{Related Work}
\label{sec:related}

\paragraph{LLMs for PDEs and Scientific Computing.}
A growing line of work treats scientific simulation as a code-generation problem and
probes the limits of LLMs in that role
\citep{li2025codepde, gaonkar2025sciml, tian2024scicode, du2026autonumerics, deotale2026all}.
Within this space, existing systems pursue different strategies. CodePDE studies PDE
solver construction directly and improves reliability through inference-time refinement
against numerical references, including reasoning chains, debugging loops, self-refinement,
and test-time scaling \citep{li2025codepde}. Other scientific-code-generation systems
emphasize domain-informed prompting, numerical-method guidance, agentic orchestration, or
fine-tuning to improve executability and numerical validity across broader scientific
computing tasks \citep{gaonkar2025sciml, du2026autonumerics, deotale2026all,
liu2025pde, wuwu2025pinnsagent}. More broadly, research-grade scientific coding benchmarks
show that realistic scientific programming remains difficult for modern LLMs, despite
strong performance on generic code benchmarks \citep{tian2024scicode}. Thus, prior work
largely improves scientific reliability through inference-time guidance, reference-solution
feedback, or agentic refinement; RLVP instead studies whether numerical and physical
preferences can be internalized through post-training.

\paragraph{LLMs as Simulator Agents and Simulator-Grounded Learners.}
A related but distinct line of work uses LLMs to operate scientific simulators or to learn
from simulation outcomes. Simulator-agent systems treat mature scientific software as an
external backend: the LLM writes input files, code wrappers, configuration scripts, or
domain-specific commands, while the numerical solver itself remains fixed by the underlying
software stack
\citep{yue2025foam, pandey2025openfoamgpt, chen2024metaopenfoam, zhang2025mooseagent,
wang2026chronollm, he2025lang, mudur2025feabench}. Other work uses simulation-derived
signals as supervision for reasoning or scientific model construction, including sparse
simulator rewards, preferences, curated physics outcomes, or execution rewards
\citep{prabhudesai2026solving, soroco2025pde, lie2026agentic}. These works establish
simulators as tools and feedback sources for LLMs. RLVP differs in the object being
trained: the model is post-trained to generate the numerical PDE solver itself, and the
verifier scores the executed solution using reference-solution error and PDE-residual
information.

\paragraph{Reinforcement Learning with Verifiable Rewards.}
RL with verifiable rewards (RLVR) has become a central post-training paradigm for code
and mathematical reasoning \citep{shao2024deepseekmath, guo2025deepseek, yu2025dapo,
zheng2025group}. In its canonical form, the verifier is binary or near-binary:
DeepSeek-R1 uses rule-based accuracy and format rewards and avoids learned reward models
in R1-Zero \citep{guo2025deepseek}, while DAPO and GSPO refine GRPO through improved
normalization, stability, and open RL system design
\citep{yu2025dapo, zheng2025group}. Related work on RL for code generation has used
unit-test outcomes, learned critic signals, compiler-feedback curricula, and iterative
execution-feedback grounding
\citep{le2022coderl, dou2024stepcoder, gehring2024rlef, yang2025reinforcement}. Recent
methods move beyond strict binary supervision through process-level reward models and
graded correctness signals \citep{damani2025beyond, uesato2022solving}. However, these
verifiers still largely operate on discrete outcomes, labels, or program-execution.

\paragraph{Scientific Machine Learning.}
A central paradigm in scientific machine learning trains neural networks to approximate
PDE solution operators or solution fields directly. Physics-informed neural networks
(PINNs) and related methods minimize PDE residuals on network outputs
\citep{raissi2019physics}; neural operators learn solution maps between function spaces
\citep{li2020fourier, lu2021learning, cai2024longrollout}; and benchmarks such as PDEBench standardize the evaluation of these surrogates against numerical references using trajectory-level $L^2$ error \citep{takamoto2022pdebench}. More recently, generative PDE surrogates have used diffusion and flow-matching objectives to model distributions over physical fields \citep{huang2024diffusionpde, bastek2024physics, utkarsh2025physics, utkarsh2025end}. These approaches use
physics to train, regularize, or constrain neural surrogates that directly output physical states. In contrast, RLVP does not train a neural network to be the simulator. It post-trains a language model to \emph{write} conventional numerical-methods code, using reference-solution error and PDE residuals as execution-grounded verifier rewards on the generated program.
\vspace{-10px}
\section{Methodology}
\label{sec:method}

We frame PDE solver synthesis as a code-generation problem and post-train language models
with a verifier that provides continuous feedback over function-valued numerical outputs.
\emph{Code is the action; physics is the verifier.} The model emits a solver program, which
is run inside an execution environment and graded by comparing the resulting trajectory to
the function-space behavior of a reference solution. This differs from standard
RLVR, where correctness is typically
determined by sparse rule-based checks such as unit-test pass/fail outcomes
\citep{le2022coderl, uesato2022solving, shao2024deepseekmath}. PDE correctness is
intrinsically graded: a generated solver may execute and return arrays of the correct shape
while still being unstable, biased, diffusive, dispersive, or inconsistent with the
underlying physical law. Our verifier exploits this graded structure to provide a richer
optimization signal than binary execution outcomes alone, while retaining the rule-based
verifiability that makes RLVR attractive.

Our pipeline has two stages. Supervised fine-tuning (SFT) on a curated bank of teacher solvers initializes the policy within the space of executable numerical programs. RL then sharpens this prior using a hybrid binary--continuous verifier that rewards execution validity, physical accuracy, and PDE residual consistency. Thus, we follow the standard warm-start SFT plus RL recipe
\citep{ouyang2022training, shao2024deepseekmath, guo2025deepseek}, but replace discrete answer verification with execution-grounded physical verification.

\subsection{Numerical PDE Solver Synthesis as a Code-Generation Task}
\label{subsec:solver_synthesis}

Let $x$ denote a prompt specifying a PDE, inputs, and the required code interface. A
policy $\pi_\theta$ generates a candidate solver program
\begin{math}
    y \sim \pi_\theta(\cdot \mid x),
\end{math}
where $y$ is an implementation, not a numerical solution array. Each program is evaluated
on hidden PDE instances
\begin{math}
    c = (\Omega_h, \mathcal{T}, u_0, \mu, b),
\end{math}
where $\Omega_h$ is the mesh, $\mathcal{T}$ the requested output times, $u_0$ the initial
condition when applicable, $\mu$ the PDE parameters, and $b$ the boundary/forcing data.
Sandboxed execution returns
\begin{math}
    \widehat{U} = \mathrm{Exec}(y,c),
\end{math}
as either a steady field or a time-indexed trajectory. The verifier compares
$\widehat{U}$ to a hidden reference solution $U^\star$; physical correctness enters
training only through rewards computed after execution.

\subsection{Multi-PDE Environment}
\label{subsec:multipde_env}

We train one policy across eight PDE families spanning hyperbolic transport, nonlinear
conservation laws, parabolic reaction--diffusion, elliptic steady-state solves, and
incompressible flow in 1D and 2D. Hidden verifier instances vary grid resolutions, output times, parameters, and initial conditions; where boundary treatments also differ across PDE problems. Thus, high reward requires an algorithmic solver rather than memorized outputs. Full equations, parameter ranges, boundary conditions, and reference-solver configurations are in Appendix~\ref{app:pdes}.
\vspace{-10px}
\paragraph{Prompt design.}
Prompts are sampled in three forms: generic, parameter-conditioned, and
parameter and initial condition conditioned. All forms specify the mathematical problem and the same solver interface, but not the numerical method. Concrete examples, sampling weights, and
per-PDE counts are provided in Appendix~\ref{app:prompt_examples}.

\subsection{Solver-Bank Supervised Fine-Tuning}
\label{subsec:sft}
\vspace{-10px}
Direct RL from a base language model is inefficient because most early generations fail
before producing meaningful physical feedback. We therefore construct a solver bank
$\mathcal{B}_p$ for each PDE family $p$, containing executable teacher implementations
that cover common numerical idioms, including finite-difference stencils, conservative
updates, CFL substepping, sparse linear solves, projection steps, and boundary-condition
handling. Detailed solver-bank tables and verification checks are in
Appendix~\ref{app:bank-organization}. We report numerical verifications of the solver bank in Appendix~\ref{app:verification}, including self-convergence and MMS checks.
 
Given prompt--solver pairs
$\{(x_i,y_i^{\mathrm{teach}})\}_{i=1}^{M}$ drawn from $\bigcup_p \mathcal{B}_p$, we minimize
the standard supervised next-token loss
\begin{math}
    \mathcal{L}_{\mathrm{SFT}}(\theta)
    =
    -
    \frac{1}{M}
    \sum_{i=1}^{M}
    \log \pi_\theta(y_i^{\mathrm{teach}}\mid x_i).
    \label{eq:sft_loss}
\end{math}
SFT provides a \emph{solver prior}: it places executable schemes within the policy support
and raises execution success to the regime where verifier feedback becomes informative.
RL then optimizes hidden-instance robustness, stability, and physics consistency.

\subsection{Verifier Design for PDE Solver Generation}
\label{subsec:verifier_design}

PDE solvers do not have a clean binary notion of correctness: a program can execute, return
the right shape, and contain only finite values while still being unstable, biased, or
inconsistent with the PDE. This creates a \emph{verifiability gap}: hard checks are needed
for program feasibility, but they discard the graded numerical quality that separates a
barely stable approximation from an accurate solver. RLVP closes this gap by gating
continuous function-space feedback with binary execution checks, so invalid programs
receive zero reward while valid programs are ranked by the physical field they return.
Let $p$ index the PDE family, and let
\begin{math}
    \mathcal{R}_p[u;\mu,b] = 0
\end{math}
denote its residual form. For time-dependent problems, this may be written as
\begin{math}
    \mathcal{R}_p[u;\mu,b]
    =
    \partial_t u + \mathcal{N}_p[u;\mu,b],
\end{math}
while for steady-state problems it may encode an elliptic operator or constraint. Let
\begin{math}
    u^\star : [0,T]\times \Omega \rightarrow \mathbb{R}^{d_u}
\end{math}
denote the reference solution on spatial domain $\Omega$, and let $\widehat{u}$ denote the
generated solution. For time-dependent problems, we measure error with the normalized
space-time $L^2$ metric

\begin{equation}
\resizebox{\linewidth}{!}{$\displaystyle
    \mathcal{L}_{\mathrm{traj}}(\widehat{u},u^\star)
    =
    \frac{
    \left\|
    \widehat{u} - u^\star
    \right\|_{L^2([0,T]\times\Omega)}
    }{
    \left\|
    u^\star
    \right\|_{L^2([0,T]\times\Omega)}
    + \epsilon
    },
    \qquad
    \left\|
    \widehat{u} - u^\star
    \right\|_{L^2([0,T]\times\Omega)}^2
    =
    \int_0^T
    \int_{\Omega}
    \left\|
    \widehat{u}(t,x) - u^\star(t,x)
    \right\|_2^2
    \, dx\,dt .
$}
\label{eq:l2_traj_error}
\end{equation}

For steady-state problems, the time integral is omitted. For vector-valued systems, the
inner norm is taken over solution channels. In implementation, this integral is
approximated by the corresponding grid-weighted discrete norm on each hidden evaluation
mesh.

This gives a uniform function-space error across scalar fields, multi-field systems,
steady-state solutions, and time-dependent trajectories. Because the verifier scores the
returned field rather than the code template, it is agnostic to the numerical method used
by the generated program.

\subsection{RLVP Reward Construction}
\label{subsec:rlvp_reward}

We instantiate the verifier in three steps: a binary feasibility gate, stochastic-tolerance
scores for continuous diagnostics, and a final multiplicative reward.

\paragraph{Validity.}
For a generated solver $y$ on hidden instance $c$, let
$v_{\mathrm{exec}}$, $v_{\mathrm{shape}}$, and $v_{\mathrm{finite}}$ indicate whether the
program executes without error, returns the required output shape, and produces finite
values. Invalid programs receive zero reward. The hard validity gate is
\begin{equation}
    V(y,c)
    =
    v_{\mathrm{exec}}(y,c)\,
    v_{\mathrm{shape}}(y,c)\,
    v_{\mathrm{finite}}(y,c).
    \label{eq:validity_gate}
\end{equation}

\paragraph{Uncertain tolerances.}
For valid programs, each nonnegative diagnostic loss $\mathcal{L}_j(y,c)$ is treated as a
binary test with an uncertain acceptance tolerance. A fixed threshold would accept when
$\mathcal{L}_j(y,c)\le\epsilon_j$, but acceptable error scales vary across PDE families,
grids, discretizations, and diagnostics. We therefore draw
$\epsilon_j\sim p_{T_j}$ and define the soft-verification factor as the expected binary
pass outcome:
\begin{equation}
    R_j(y,c)
    =
    \mathbb{E}_{\epsilon_j\sim p_{T_j}}
    \left[
    \mathbf{1}\{\mathcal{L}_j(y,c)\le \epsilon_j\}
    \right]
    =
    \mathbb{P}_{\epsilon_j\sim p_{T_j}}
    \left(
    \epsilon_j \ge \mathcal{L}_j(y,c)
    \right).
    \label{eq:expected_pass_reward}
\end{equation}
Thus, the continuous reward is still an RLVR-style pass probability, now marginalizing over
uncertainty in the tolerance. With an exponential tolerance distribution this becomes
\begin{math}
    R_j(y,c)
    =
    \exp\left(
    -\frac{\mathcal{L}_j(y,c)}{T_j}
    \right).
    \label{eq:exp_expected_pass}
\end{math}
We provide the derivation, interpretation of the exponential model, and relation to Gibbs
forms in \Cref{app:expected_pass_reward}.

\paragraph{Physical accuracy and physics residual factors.}
Applying \Cref{eq:exp_expected_pass} to the function-space error from \Cref{eq:l2_traj_error} gives
\begin{math}
    R_{\mathrm{traj}}(y,c)
    =
    \exp\left(
    -\frac{
    \mathcal{L}_{\mathrm{traj}}(\widehat{u},u^\star)
    }{
    T_{\mathrm{traj}}
    }
    \right).
    \label{eq:traj_reward}
\end{math}
We refer to this as the physical accuracy reward: this factor ranks valid programs by the accuracy of the physical field or trajectory they produce on the hidden verifier instance. When a residual operator is available, we also include a physics residual consistency diagnostic. With the time integral omitted for steady-state problems, define
\begin{equation}
    \rho(\widehat{u})
    =
    \left\|
    \mathcal{R}_p[\widehat{u};\mu,b]
    \right\|_{L^2([0,T]\times\Omega)}
    \qquad
    \mathcal{L}_{\mathrm{phys}}(\widehat{u},u^\star)
    =
    \frac{
    \left|
    \rho(\widehat{u}) - \rho(u^\star)
    \right|
    }{
    \rho(u^\star) + \epsilon
    } .
    \label{eq:phys_loss}
\end{equation}
Comparing to $\rho(u^\star)$ absorbs the discretization error of the reference solution
and avoids rewarding residuals smaller than the numerical reference itself. The
corresponding expected-pass factor is
\begin{math}
    R_{\mathrm{phys}}(y,c)
    =
    \exp\left(
    -\frac{
    \mathcal{L}_{\mathrm{phys}}(\widehat{u},u^\star)
    }{
    T_{\mathrm{phys}}
    }
    \right)
    \label{eq:phys_reward}
\end{math}. Finally, the complete RLVP reward is
\begin{equation}
    R(y,c)
    =
    V(y,c)
    \cdot
    R_{\mathrm{traj}}(y,c)
    \cdot
    R_{\mathrm{phys}}(y,c).
    \label{eq:final_reward}
\end{equation}
If no residual diagnostic is used, we set $R_{\mathrm{phys}}=1$.
The residual term also acts as an attenuation factor rather than an additive auxiliary
objective: it penalizes physically inconsistent returned solutions without introducing
reward for surface-level code features.

\subsection{RL Post-Training with GRPO}
\label{subsec:grpo}

Starting from the SFT-initialized policy, we apply Group-Relative Policy Optimization
(GRPO) \citep{shao2024deepseekmath}. For each prompt $x_i$, we sample a group of $G$
candidate programs from the old policy,
\begin{math}
    y_{i,1},\ldots,y_{i,G}
    \sim
    \pi_{\theta_{\mathrm{old}}}(\cdot\mid x_i),
\end{math}
execute each program in the PDE environment, and compute verifier rewards
$R_{i,g}=R(y_{i,g},x_i)$. Following GRPO, advantages are computed from the raw verifier
rewards, without KL shaping, by normalizing rewards within each group:
\begin{math}
    A_{i,g}
    =
    \frac{
    R_{i,g}
    -
    \operatorname{mean}_{g'} R_{i,g'}
    }{
    \operatorname{std}_{g'} R_{i,g'}+\epsilon
    }.
    \label{eq:grpo_advantage}
\end{math}
The same output-level advantage $A_{i,g}$ is assigned to each valid response token in the
generated program $y_{i,g}$. Let
\begin{math}
    \rho_{i,g,t}(\theta)
    =
    \frac{
    \pi_\theta(y_{i,g,t}\mid x_i,y_{i,g,<t})
    }{
    \pi_{\theta_{\mathrm{old}}}(y_{i,g,t}\mid x_i,y_{i,g,<t})
    }
    \label{eq:grpo_ratio}
\end{math}
denote the token-level policy ratio. The clipped GRPO objective is

\begin{equation}
\resizebox{\linewidth}{!}{$\displaystyle
    \mathcal{J}_{\mathrm{GRPO}}(\theta)
    =
    \mathbb{E}_{i}
    \left[
    \frac{1}{G}
    \sum_{g=1}^{G}
    \frac{1}{|y_{i,g}|}
    \sum_{t=1}^{|y_{i,g}|}
    \left\{
    \min\left(
    \rho_{i,g,t}(\theta) A_{i,g},
    \operatorname{clip}
    \left(
    \rho_{i,g,t}(\theta),
    1-\epsilon_{\mathrm{clip}},
    1+\epsilon_{\mathrm{clip}}
    \right)
    A_{i,g}
    \right)
    -
    \beta_{\mathrm{KL}}\,
    \mathbb{D}_{\mathrm{KL}}
    \!\left[
    \pi_\theta \,\|\, \pi_{\mathrm{ref}}
    \right]_{i,g,t}
    \right\}
    \right].
$}
\label{eq:grpo_objective}
\end{equation}
where
$\mathbb{D}_{\mathrm{KL}}
    \!\left[
    \pi_\theta \,\|\, \pi_{\mathrm{ref}}
    \right]_{i,g,t}$
is the per-token KL penalty to the frozen SFT reference model $\pi_{\mathrm{ref}}$. In our
implementation, we use the standard low-variance sample-based approximation
\begin{equation}
    \mathbb{D}_{\mathrm{KL}}
    \!\left[
    \pi_\theta \,\|\, \pi_{\mathrm{ref}}
    \right]_{i,g,t}
    =
    q_{i,g,t}
    -
    \log q_{i,g,t}
    -
    1,
    \qquad
    q_{i,g,t}
    =
    \frac{
    \pi_{\mathrm{ref}}(y_{i,g,t}\mid x_i,y_{i,g,<t})
    }{
    \pi_{\theta}(y_{i,g,t}\mid x_i,y_{i,g,<t})
    }.
    \label{eq:kl_approx}
\end{equation}
Here $\pi_{\theta_{\mathrm{old}}}$ is the policy used to sample the current batch, while
$\pi_{\mathrm{ref}}$ is used only for KL regularization. The group baseline removes the need for a learned value model, and the verifier rewards shift probability mass toward programs that satisfy hidden execution checks and produce physically accurate solutions. Our implementation uses token-mean aggregation over valid response tokens. We examine the resulting training dynamics: reward trajectories, and nRMSE trends in \Cref{sec:experiments}.

\vspace{-10px}
\section{Empirical Evaluation}
\label{sec:experiments}

\vspace{-5px}
\subsection{Experimental Setup}
\label{subsec:exp_setup}
\paragraph{Training data and prompts.}
We post-train on eight seen PDE families: 1D advection, 1D Burgers, 1D reaction--diffusion, 1D diffusion--sorption, 2D reaction--diffusion, 2D Darcy flow, 2D incompressible Navier--Stokes, and 2D shallow water. These tasks span hyperbolic transport, parabolic diffusion and reaction, elliptic solves, and coupled 2D conservation-law dynamics. During RLVP, each prompt is paired with hidden PDE instances (varied by initial conditions and PDE parameters) sampled from the training environment, so the policy is optimized against executable function-space accuracy rather than token imitation alone. Full PDE specifications, parameter grids, boundary and initial conditions, and solver bank coverage are provided in Appendix~\ref{app:pde-bank}. Hyperparameters and compute details are in Appendix~\ref{app:exp-details}.

\vspace{-20px}
\paragraph{Evaluation problems and metrics.}
We evaluate all models on fixed PDE prompts and case sets. The seen-PDE evaluation uses the same eight PDE families as training, with 143 total cases. The unseen PDE evaluation uses 10 held-out PDE prompts and 137 cases: 2D advection, 1D and 2D advection--diffusion, 1D advection--reaction--diffusion, 2D Allen--Cahn, 1D Cahn--Hilliard, 2D Darcy--reaction, 2D Gray--Scott, 1D heat, and 1D KdV. These held-out PDEs are not used as SFT solver-bank targets or as RL prompts. For each checkpoint, the model generates \(k=8\) solver programs per PDE prompt, and each program is evaluated on the fixed test cases for that prompt. We report combinatorial pass@\(k\), valid execution rate, and median best-of-8 nRMSE as the continuous physical accuracy metric. Detailed validity criteria, metric definitions, evaluator fallbacks, and aggregation procedures are provided in Appendix~\ref{app:evaluation-details}.

\subsection{RLVP Improves Numerical PDE Solver Code Generation}
\label{subsec:main_results}
\begin{figure}[t!]
    \centering
    \includegraphics[width=0.85\linewidth]{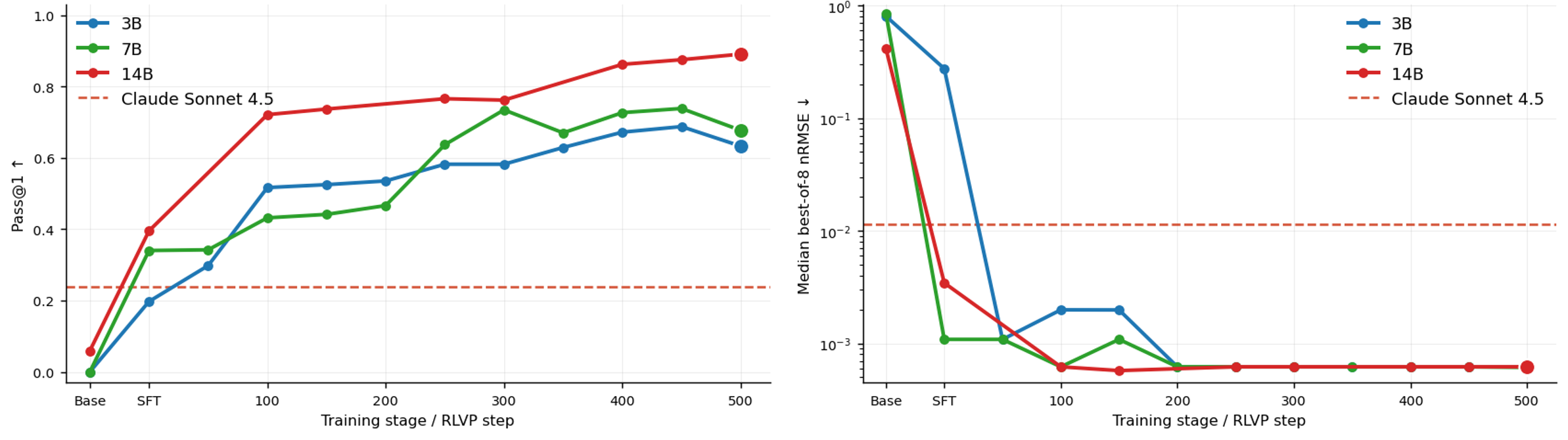}
    \caption{
    \textbf{Progress of evaluation metrics during post-training.}
    We compare Base (Qwen2.5-Coder-Instruct), SFT, and RLVP post-trained checkpoints on the eight seen PDE tasks. Dashed line: Claude Sonnet 4.5 baseline evaluated with the same \(k=8\) sampling protocol.
    }
    \label{fig:id-progress}
\end{figure}

We first quantify how the evaluation metrics change after verifier-based post-training on the eight seen PDE prompts. SFT teaches the solver program prior: function signatures, correct array shapes, handling of PDE-specific parameters and boundary conditions, and common numerical solver templates. RLVP then sharpens this prior with execution-grounded feedback, reweighting the prior towards generating solver programs that both execute and solve a broad range of PDE problems with hidden cases of varying discretizations, time grids, parameters, and initial conditions. As shown in Figure~\ref{fig:id-progress}, RLVP improves both single-sample pass@1 reliability and best-of-8 accuracy across the 3B, 7B, and 14B Qwen2.5-Coder-Instruct models. The corresponding training reward curves are shown in Figure~\ref{fig:training-reward-curves}.

\begin{figure}[H]
    \centering
    \includegraphics[width=1\linewidth]{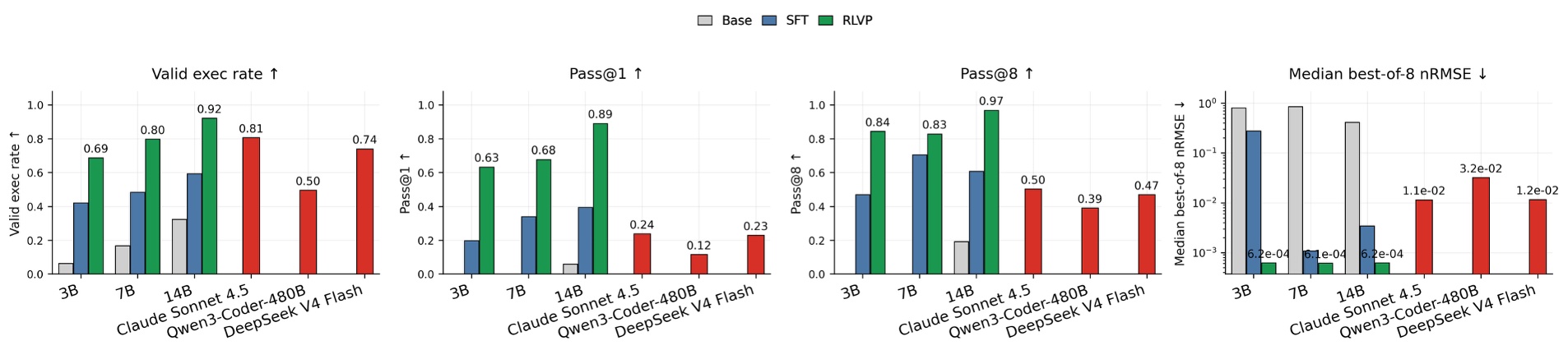}
    \caption{\textbf{Comparison across local and frontier models on seen PDEs.} Base, SFT, RLVP, and frontier LLMs are compared by valid execution rate, pass@1, pass@8, and median best-of-8 nRMSE.}
    \label{fig:id-final-bars}
\end{figure}

Table~\ref{tab:id-main} and Figure~\ref{fig:id-final-bars} summarize the aggregate and per-PDE performance on the seen tasks. Across all three model scales, RLVP raises pass@1 and pass@8 from near-zero base model performance to \(0.63\)--\(0.89\) and \(0.83\)--\(0.97\) respectively and drives median-best nRMSE@8 to approximately \(6\times10^{-4}\). The post-trained local models also outperform static frontier model baselines under the same direct-generation protocol. For example, the Qwen2.5-Coder-Instruct model (7B) post-trained with RLVP reaches pass@1/pass@8 \(0.68/0.83\) and nRMSE@8 \(6.13\times10^{-4}\), compared with \(0.24/0.50\) and \(1.15\times10^{-2}\) for Claude Sonnet 4.5. 

\vspace{-5px}
With RLVP, parameter updates from executable physical feedback can make a smaller policy more likely to generate numerically reliable solver programs than prompting a much larger static model. Our multi-PDE setup enables all models to solve most PDE tasks simultaneously and robustly after RLVP. Incompressible Navier--Stokes remains the hardest PDE task as it requires coupled velocity evolution and projection-like structure. The 14B RLVP checkpoint is the only local model with strong Navier--Stokes solver generation (Table~\ref{tab:id-main}), reaching pass@1 \(0.62\) and nRMSE@8 \(3.48\times10^{-3}\), suggesting that this PDE benefits from a larger model capacity.

\begin{table*}[t]
\centering
\scriptsize
\caption{
\textbf{Aggregate and per-PDE evaluation performance.}
We report aggregate metrics together with per-PDE pass@1 and median best-of-8 nRMSE. Additional metrics are provided in Appendix Table~\ref{tab:id-appendix}. Best and second-best values in each row are marked in \textbf{bold} and \underline{underline}.
}
\label{tab:id-main}
\resizebox{\textwidth}{!}{%
\begin{tabular}{llcccccccccccc}
\toprule
 &  & \multicolumn{2}{c}{3B} & \multicolumn{2}{c}{7B} & \multicolumn{2}{c}{14B} & \multicolumn{1}{c}{Claude Sonnet 4.5} & \multicolumn{1}{c}{Qwen3-Coder-480B} & \multicolumn{1}{c}{DeepSeek V4 Flash} & \multicolumn{1}{c}{GPT-OSS-120B} & \multicolumn{1}{c}{Llama-3.1-405B} & \multicolumn{1}{c}{Llama-3.3-70B} \\
\cmidrule(lr){3-4} \cmidrule(lr){5-6} \cmidrule(lr){7-8} \cmidrule(lr){9-9} \cmidrule(lr){10-10} \cmidrule(lr){11-11} \cmidrule(lr){12-12} \cmidrule(lr){13-13} \cmidrule(lr){14-14}
Section & Metric & Base & RLVP & Base & RLVP & Base & RLVP &  &  &  &  &  &  \\
\midrule
Aggregate & Pass@1 $\uparrow$ & 0.00 & 0.63 & 0.00 & \underline{0.68} & 0.06 & \textbf{0.89} & 0.24 & 0.12 & 0.23 & 0.21 & 0.07 & 0.04 \\
 & Pass@8 $\uparrow$ & 0.00 & \underline{0.84} & 0.00 & 0.83 & 0.19 & \textbf{0.97} & 0.50 & 0.39 & 0.47 & 0.43 & 0.29 & 0.19 \\
 & Median-best nRMSE@8 $\downarrow$ & 8.03E-01 & \underline{6.20E-04} & 8.46E-01 & \textbf{6.13E-04} & 4.14E-01 & 6.20E-04 & 1.15E-02 & 3.22E-02 & 1.17E-02 & 1.22E-02 & 1.67E-01 & 7.99E-01 \\
 & Valid exec rate $\uparrow$ & 0.06 & 0.69 & 0.17 & 0.80 & 0.32 & \textbf{0.92} & \underline{0.81} & 0.50 & 0.74 & 0.58 & 0.27 & 0.21 \\
\midrule
Per-PDE Pass@1 & Advection 1D & 0.00 & \textbf{1.00} & 0.00 & \underline{0.88} & 0.04 & 0.75 & 0.14 & 0.43 & 0.38 & 0.38 & 0.00 & 0.00 \\
 & Burgers 1D & 0.00 & \underline{0.69} & 0.00 & \textbf{0.75} & 0.00 & \textbf{0.75} & 0.25 & 0.08 & 0.12 & 0.12 & 0.12 & 0.00 \\
 & Darcy Flow 2D & 0.00 & \underline{0.75} & 0.00 & 0.67 & 0.00 & \textbf{1.00} & 0.00 & 0.00 & 0.00 & 0.00 & 0.00 & 0.00 \\
 & Diffusion-sorption 1D & 0.00 & \underline{0.88} & 0.00 & \underline{0.88} & 0.03 & \textbf{1.00} & 0.00 & 0.00 & 0.00 & 0.00 & 0.00 & 0.03 \\
 & Incompressible Navier Stokes 2D & 0.00 & 0.00 & 0.00 & 0.00 & 0.00 & \textbf{0.62} & 0.04 & 0.00 & 0.00 & \underline{0.12} & 0.00 & 0.00 \\
 & Reaction Diffusion 1D & 0.00 & \underline{0.88} & 0.00 & \textbf{1.00} & 0.41 & \textbf{1.00} & 0.59 & 0.17 & 0.87 & 0.00 & 0.25 & 0.25 \\
 & Reaction Diffusion 2D & 0.00 & 0.75 & 0.00 & 0.88 & 0.00 & \textbf{1.00} & 0.50 & 0.25 & 0.25 & \underline{0.92} & 0.15 & 0.04 \\
 & Shallow Water 2D & 0.00 & 0.12 & 0.00 & \underline{0.38} & 0.00 & \textbf{1.00} & \underline{0.38} & 0.00 & 0.21 & 0.12 & 0.00 & 0.00 \\
\midrule
Per-PDE nRMSE@8 & Advection 1D & 6.05E-01 & \underline{9.44E-04} & 2.51E-01 & \textbf{9.29E-04} & 2.86E-02 & \underline{9.44E-04} & 1.34E-02 & 1.33E-02 & 1.32E-02 & 1.34E-02 & 3.27E-01 & 1.00E+00 \\
 & Burgers 1D & 1.00E+00 & 3.03E-03 & 1.00E+00 & 3.02E-03 & 2.29E-01 & 3.03E-03 & \underline{1.78E-03} & 4.64E-03 & \textbf{1.51E-03} & 1.10E-02 & 4.03E-02 & 1.00E+00 \\
 & Darcy Flow 2D & 1.00E+00 & \textbf{3.85E-15} & 1.00E+00 & \textbf{3.85E-15} & 5.98E-01 & \textbf{0.00E+00} & 1.08E-01 & 4.86E-01 & \underline{3.56E-02} & 1.08E-01 & 2.79E-01 & 5.98E-01 \\
 & Diffusion-sorption 1D & 1.00E+00 & \underline{2.97E-04} & 3.26E-01 & \textbf{2.96E-04} & 3.56E-02 & 2.97E-04 & 5.35E-02 & 3.04E-01 & 5.07E-02 & 5.35E-02 & 1.00E+00 & 3.56E-02 \\
 & Incompressible Navier Stokes 2D & 1.00E+00 & 1.00E+00 & 1.00E+00 & 3.33E-01 & 1.00E+00 & \textbf{3.48E-03} & 1.67E-02 & 1.01E-01 & 2.78E-02 & \underline{6.45E-03} & 1.00E+00 & 1.00E+00 \\
 & Reaction Diffusion 1D & 2.98E-01 & 3.71E-06 & 6.92E-01 & 3.71E-06 & 3.18E-04 & \textbf{9.73E-08} & 1.61E-05 & 4.52E-05 & 3.71E-06 & 1.00E+00 & \underline{3.70E-06} & 3.70E-06 \\
 & Reaction Diffusion 2D & 1.60E-01 & \underline{1.11E-05} & 1.51E-01 & \underline{1.11E-05} & 1.00E+00 & \textbf{8.44E-06} & 4.67E-03 & 3.90E-03 & 4.03E-03 & 3.90E-03 & 8.89E-03 & 4.86E-02 \\
 & Shallow Water 2D & 5.11E-02 & \underline{1.23E-03} & 1.00E+00 & 1.23E-03 & 1.00E+00 & \textbf{1.14E-03} & 9.56E-03 & 5.11E-02 & 1.02E-02 & 1.02E-02 & 5.53E-02 & 1.00E+00 \\
\bottomrule
\end{tabular}%
}
\end{table*}

\vspace{-5px}
\subsection{Compositional Transfer to Unseen PDE Families}
\label{subsec:ood}

We next evaluate whether the learned prior transfers beyond the PDE families used for SFT and RL. The 10 held-out tasks cover diffusion, 2D advection, advection--reaction--diffusion, Cahn--Hilliard, Gray--Scott, and Korteweg–De Vries (KdV) equations. This setting should be interpreted as a test of selective numerical-methods transfer, rather than as evidence that the model can solve arbitrary unseen PDEs.

\begin{figure}[H]
    \centering
    \includegraphics[width=1\linewidth]{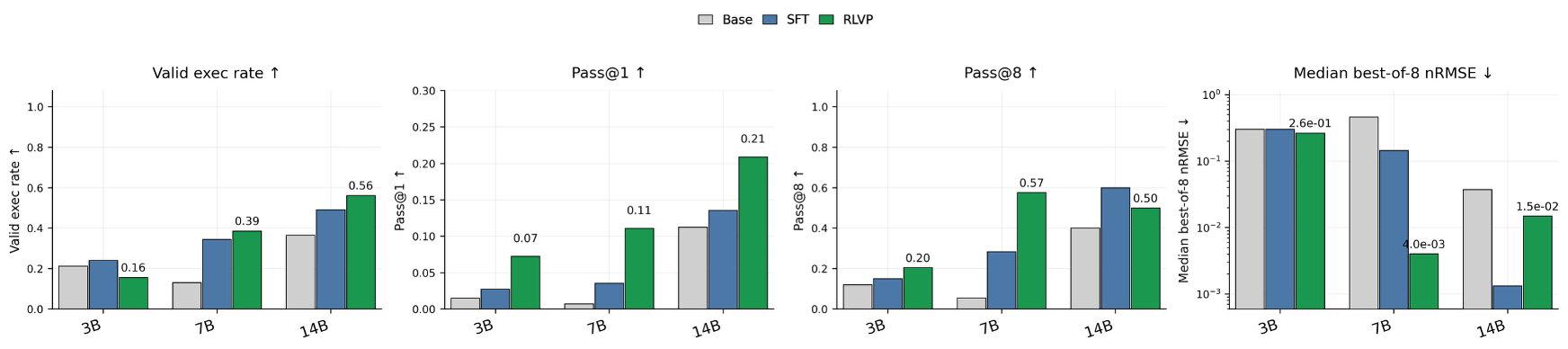}
    \caption{
    \textbf{Evaluation on 10 held-out PDE tasks.}
    We compare Base, SFT, and RLVP checkpoints on unseen PDE prompts not used during SFT or RL training.
    }
    \label{fig:ood-bars}
\end{figure}

Figure~\ref{fig:ood-bars} reports aggregate performance on the unseen PDE problems. Although these PDEs are absent from both the SFT solver bank and the RL prompt distribution, RLVP improves pass@1 relative to the base and SFT checkpoints across model sizes. For example in the 7B model, pass@8 increases from \(0.28\) after SFT to \(0.57\) after RLVP, while median-best nRMSE@8 decreases from \(1.45\times10^{-1}\) to \(4.01\times10^{-3}\). Since the held-out prompts were not training objectives, these results are best viewed as evidence that RLVP increases the probability of the LLM discovering reusable solver structure on related PDEs. Aggregate metrics and per-PDE breakdowns are provided in Appendix Tables~\ref{tab:ood-aggregate-app} and~\ref{tab:ood-appendix}.

The per-PDE breakdown in Appendix Table~\ref{tab:ood-appendix} shows that the 7B improvement is not driven by a single held-out equation. RLVP is strongest on heat, 2D advection, advection--reaction--diffusion, and Cahn--Hilliard. For 7B, advection--reaction--diffusion improves from SFT pass@8 \(0.00\) to RLVP pass@8 \(1.00\), with nRMSE@8 dropping from \(5.25\times10^{-1}\) to \(1.50\times10^{-3}\). Heat similarly improves from SFT pass@8 \(0.00\) to RLVP pass@8 \(1.00\), with nRMSE@8 \(8.11\times10^{-5}\). In contrast, advection--diffusion, Darcy--reaction, and KdV equations remain difficult, indicating that transfer is selective and depends on the numerical structures available in the trained policy. We examine advection--reaction--diffusion below as a representative case of compositional numerical methods transfer.

\begin{figure}
    \centering
    \includegraphics[width=0.9\linewidth]{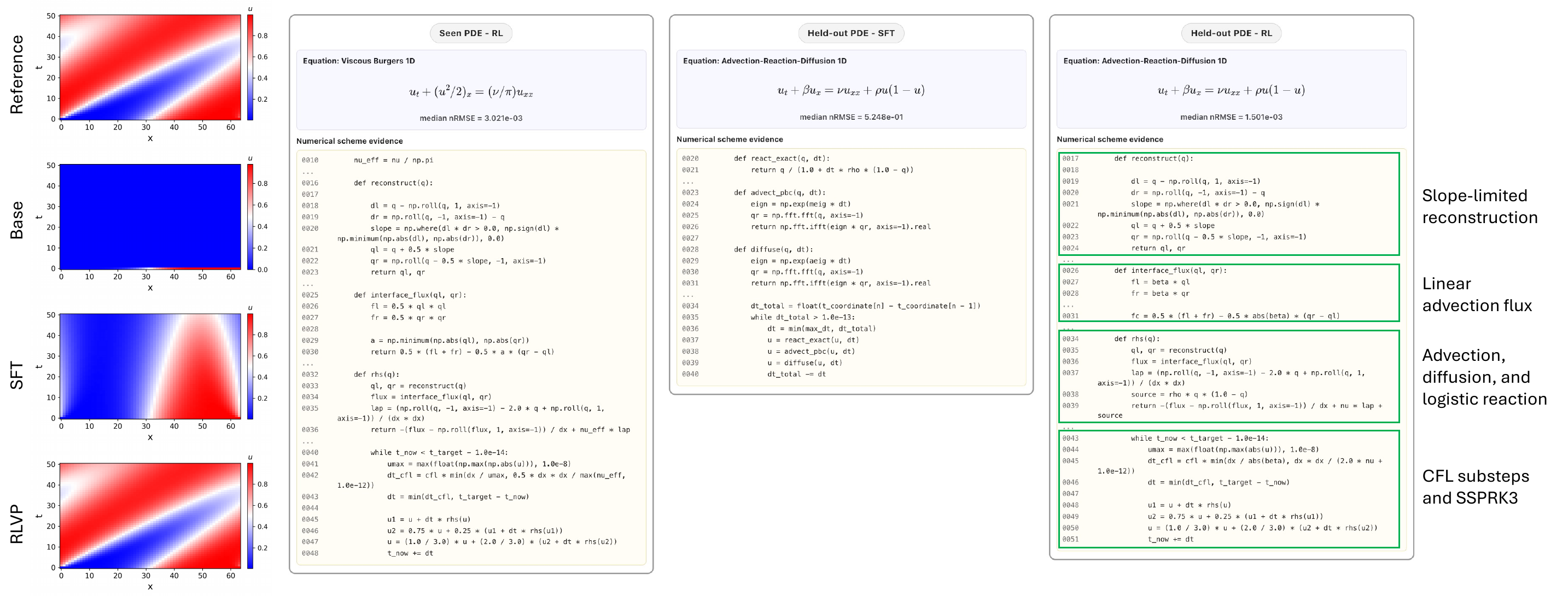}
    \caption{
    \textbf{Compositional transfer to unseen advection--reaction--diffusion (ARD)} (RLVP 7B model).
    Left: reference and best-of-8 generated solutions for the held-out ARD problem with \(\beta=1\), \(\nu=0.02\), and \(\rho=2\). Right: numerical scheme evidences from generated PDE solver snippets: a Burgers solver from RLVP, and ARD solvers from SFT and RLVP post-trained models. 
    }
    \label{fig:transfer-ard}
\end{figure}

\paragraph{Evidence of compositional numerical methods transfer.}
Figure~\ref{fig:transfer-ard} examines advection--reaction--diffusion (ARD), a held-out PDE that combines linear advection, diffusion, and logistic reaction. The base model generates an almost trivial solution and the SFT model gives a qualitatively incorrect trajectory. The RLVP post-trained model generates a solver which gives a solution that closely matches the reference. The code comparison suggests that the generated solver from RLVP uses a finite-volume/stencil motif learned on the seen transport PDEs (such as the generated solver for Burgers' equation), but adapts it to the new operator composition by combining advection, reaction, and diffusion terms. Rather than directly copying an SFT template, the post-trained model appears to exhibit scheme-level compositional reuse.

Additional results on unseen PDE tasks are shown in Appendix~\ref{app:transfer-examples}. The heat equation example illustrates a diffusion-only subset of the seen reaction--diffusion family, with median nRMSE improving from \(2.07\times10^{-1}\) (SFT) to \(8.11\times10^{-5}\). The unseen 2D advection task shows dimensional lifting from seen 1D advection, extending the transport update to two spatial dimensions with CFL control and RK4 integration. Cahn--Hilliard illustrates higher-order operator synthesis by constructing the chemical potential and applying a second Laplacian. Together with the aggregate held-out results, these examples suggest that an LLM post-trained with RLVP can reuse and adapt numerical solver motifs on unseen PDE tasks, while such transfer is selective rather than universal. 

Solver components analysis gives a complementary view of the learned distributions: RLVP post-trained checkpoints more often emit \texttt{np.roll}-based stencils, CFL/substep control, and RK time-stepping schemes, consistent with a reusable and robust methods-of-lines pattern (Figure~\ref{fig:feature-frequency}).

\subsection{Physical Accuracy Improves Beyond Validity}
\label{subsec:reward-ablation}

Appendix Figures~\ref{fig:id_validity_nrmse_ablation} and~\ref{fig:ood_validity_nrmse_ablation} isolate the effect of adding the continuous physical accuracy reward \(R_{\mathrm{traj}}\), computed from function-space nRMSE, to validity-only RL for the 7B model (comparing \(R=V\) against \(R=V R_{\mathrm{traj}}\)). On seen PDEs, validity-only RL is worse on pass@1 (\(0.58\) vs.\ \(0.71\)), pass@8 (\(0.76\) vs.\ \(0.84\)), and median best-of-8 nRMSE (\(1.0\times10^{-3}\) vs.\ \(6.2\times10^{-4}\)). On unseen PDEs, adding function-space accuracy improves valid execution (\(0.45\) vs.\ \(0.29\)), pass@1 (\(0.10\) vs.\ \(0.03\)), pass@8 (\(0.33\) vs.\ \(0.21\)), and nRMSE@8 (\(1.3\times10^{-1}\) vs.\ \(1.8\times10^{-1}\)). These results indicate that binary reward (validity) alone can reward program executability, but does not provide enough information to refine physical solution quality; the dense physical-accuracy term is needed to refine solver quality and support numerical methods transfer, consistent with the verifier design in~\Cref{subsec:rlvp_reward}.

\vspace{-10px}
\section{Conclusion}
\label{sec:conclusion}
\vspace{-5px}
We have shown that adapting language models with execution-grounded RL can substantially
improve PDE solver generation. Starting from a solver-bank SFT warm start, RLVP uses a
verifiable physics reward to move beyond executability and optimize for physical accuracy and residual consistency. Across model sizes, this post-training improves
one-shot reliability, pass@$k$, and best-of-$k$ numerical accuracy, and under the same
direct-generation protocol allows smaller verifier-trained models to match or exceed much
larger static prompting baselines. We also find evidence of selective cross-family transfer:
the trained policy reuses numerical motifs such as CFL substepping, finite-difference
stencils, boundary handling, and stability safeguards on PDEs absent from training.

These findings suggest that future solver-writing LLMs should place more emphasis on
adapting model parameters with verifiable physical feedback, rather than relying only on
larger static models or inference-time scaffolding. Our study covers eight structured PDE families, leaving broader scientific settings such as
complex geometries, richer boundary conditions, adaptive meshes, stochastic PDEs, and
multiphysics systems to future work. Further efforts can  combine RLVP with retrieval, debugging loops, verifier-guided search, or
test-time scaling, instead of direct-generation
protocol. Nevertheless, this work is the first to show what post-training can
achieve for multi-PDE solver generation even with compact models and limited compute. We
expect the approach to strengthen with broader PDE coverage, richer solver banks, stronger
base models, and extensions to other scientific code-generation settings such as inverse
problems, PDE-constrained optimization, finite-element simulation, and molecular dynamics.

\section*{Acknowledgments}

P.C. is supported by the Regenerative Energy-Efficient Manufacturing of Thermoset Polymeric Materials (REMAT) Energy Frontier Research Center, funded by the U.S. Department of Energy, Office of Science, Basic Energy Sciences under award DE-SC0023457. This research used resources of the National Energy Research Scientific Computing Center (NERSC), a U.S. Department of Energy Office of Science User Facility, using NERSC award ALCC-ERCAP 0038200. We would like to thank Branden Morioka and Florencia Ojeda for early efforts.

This material is based upon work supported by the U.S. National Science Foundation under award Nos CNS-2346520,  RISE-2425761, and DMS-2325184, by the Defense Advanced Research Projects Agency (DARPA) under Agreement No. HR00112490488,  by the Department of Energy, National Nuclear Security Administration under Award Number DE-NA0004266 and by the United States Air Force Research Laboratory under Cooperative Agreement Number FA8750-19-2-1000.  Neither the United States Government nor any agency thereof, nor any of their employees, makes any warranty, express or implied, or assumes any legal liability or responsibility for the accuracy, completeness, or usefulness of any information, apparatus, product, or process disclosed, or represents that its use would not infringe privately owned rights. Reference herein to any specific commercial product, process, or service by trade name, trademark, manufacturer, or otherwise does not necessarily constitute or imply its endorsement, recommendation, or favoring by the United States Government or any agency thereof. The views and opinions of authors expressed herein do not necessarily state or reflect those of the United States Government or any agency thereof.




\newpage
\bibliography{refs}
\bibliographystyle{unsrtnat}

\newpage
\appendix
\section{Experimental Details}
\label{app:exp-details}

\Cref{tab:model-size-hyperparams} reports the hyperparameters used in post-training across model sizes.

\subsection{Training Configuration}
\label{subsec:train_config}

Training proceeds in two stages on the same Qwen2.5-Coder-Instruct backbone~\citep{hui2024qwen2}: an SFT warm-start of one epoch over $2048$ teacher examples of solver code with LoRA rank $32$~\citep{hu2022lora}, followed by GRPO~\citep{shao2024deepseekmath} implemented via verl~\citep{sheng2025hybridflow} on $512$ training problem instances for $50$ epochs with rank-$32$ LoRA. GRPO uses $8$ rollouts per prompt, a fixed KL coefficient of $10^{-3}$ to a frozen reference policy (i.e. SFT model), learning rate of $2 \times 10^{-6}$, and PPO mini-batch size $16$. Rollouts are sampled with vLLM~\citep{kwon2023efficient} at temperature $0.7$ and top-$p = 1.0$ with \texttt{max\_response\_length} $3072$. We use an AdamW optimizer and the SFT phase uses a constant learning rate.

The complete RLVP reward for a rollout follows \Cref{eq:final_reward}:
\(r = V(y,c)R_{\mathrm{traj}}(y,c)R_{\mathrm{phys}}(y,c)\). The validity gate
\(V(y,c)\) is \textbf{1} when the generated code executes and returns a finite, correctly-shaped solution field or trajectory, and zero otherwise. In implementation,
\(\mathcal{L}_{\mathrm{traj}}\) is the rollout nRMSE, so
\(R_{\mathrm{traj}}=\exp(-\mathrm{nRMSE}/T_{\mathrm{traj}})\) with
\(T_{\mathrm{traj}}=0.05\). The physics residual consistency term uses
the reference-relative residual loss from \Cref{eq:phys_loss}, giving
\(R_{\mathrm{phys}}=\exp(-\mathcal{L}_{\mathrm{phys}}/T_{\mathrm{phys}})\) with
\(T_{\mathrm{phys}}=2.0\). Separately, a rollout is counted as a \emph{success} for evaluation when $\mathrm{nRMSE} \le 10^{-2}$, this threshold is used for evaluations in pass@$k$. The training, validation, and test splits contain $512$, $64$, and $128$ PDE cases respectively, drawn approximately uniformly across the eight PDE families using a fixed seed of $1234$, and with disjoint PDE-specific sets of parameters and initial condition combinations (refer to ~\ref{app:pdes}) across splits.

\begin{table}
\centering
\scriptsize
\caption{Hyperparameters for the 3B, 7B, and 14B model RLVP post-training runs.}
\label{tab:model-size-hyperparams}
\resizebox{\linewidth}{!}{
\begin{tabular}{@{}llll@{}}
\toprule
Hyperparameter & 3B & 7B & 14B \\
\midrule
\texttt{model} & Qwen2.5-Coder-3B-Instruct & Qwen2.5-Coder-7B-Instruct & Qwen2.5-Coder-14B-Instruct \\
\texttt{active\_pdes} & 8 & 8 & 8 \\
\midrule
\texttt{max\_prompt\_length} & 1024 & 1024 & 1024 \\
\texttt{max\_response\_length} & 3072 & 3072 & 3072 \\
\texttt{train\_batch\_size} & 48 & 48 & 48 \\
\texttt{total\_epochs} & 50 & 50 & 50 \\
\texttt{train\_rows} & 512 & 512 & 512 \\
\texttt{val\_rows} & 64 & 64 & 64 \\
\texttt{test\_rows} & 128 & 128 & 128 \\
\texttt{nnodes} & 1 & 1 & 1 \\
\texttt{n\_gpus\_per\_node} & 8 & 8 & 8 \\
\midrule
\texttt{sft\_num\_examples} & 2048 & 2048 & 2048 \\
\texttt{sft\_max\_len} & 4096 & 4096 & 4096 \\
\texttt{sft\_batch\_size} & 16 & 16 & 16 \\
\texttt{sft\_epochs} & 1 & 1 & 1 \\
\texttt{sft\_lr} & $2\times10^{-5}$ & $2\times10^{-5}$ & $2\times10^{-5}$ \\
\texttt{sft\_lora\_rank} & 32 & 32 & 32 \\
\texttt{sft\_lora\_alpha} & 32 & 32 & 32 \\
\midrule
\texttt{grpo\_lora\_rank} & 32 & 32 & 32 \\
\texttt{grpo\_lora\_alpha} & 32 & 32 & 32 \\
\texttt{actor\_lr} & $2\times10^{-6}$ & $2\times10^{-6}$ & $2\times10^{-6}$ \\
\texttt{actor\_ppo\_mini\_batch\_size} & 16 & 16 & 16 \\
\texttt{actor\_ppo\_micro\_batch\_size\_per\_gpu} & 4 & 4 & 4 \\
\texttt{kl\_loss\_coef} & 0.001 & 0.001 & 0.001 \\
\texttt{rollout\_n} & 8 & 8 & 8 \\
\texttt{rollout\_temperature} & 0.7 & 0.7 & 0.7 \\
\texttt{rollout\_top\_p} & 1.0 & 1.0 & 1.0 \\
\texttt{rollout\_tensor\_parallel\_size} & 1 & 1 & 1 \\
\texttt{rollout\_max\_model\_len} & 4096 & 4096 & 4096 \\
\texttt{rollout\_max\_num\_batched\_tokens} & 49152 & 49152 & 32768 \\
\texttt{rollout\_max\_num\_seqs} & 128 & 128 & 96 \\
\texttt{vllm\_gpu\_memory\_utilization} & 0.8 & 0.8 & 0.75 \\
\texttt{rollout\_log\_prob\_micro\_batch\_size\_per\_gpu} & 8 & 8 & 4 \\
\texttt{ref\_log\_prob\_micro\_batch\_size\_per\_gpu} & 8 & 8 & 4 \\
\texttt{reward\_num\_workers} & 16 & 16 & 16 \\
\texttt{success\_nrmse\_threshold} & 0.01 & 0.01 & 0.01 \\
\midrule
\texttt{passk\_k\_values} & 1, 4, 8 & 1, 4, 8 & 1, 4, 8 \\
\texttt{passk\_temperature} & 0.7 & 0.7 & 0.7 \\
\texttt{passk\_top\_p} & 1.0 & 1.0 & 1.0 \\
\texttt{passk\_max\_new\_tokens} & 3072 & 3072 & 3072 \\
\texttt{passk\_sample\_seed} & 2026 & 2026 & 2026 \\
\bottomrule
\end{tabular}
}
\end{table}

\subsection{Computational Resources}
\label{subsec:comp_res}

All training ran on a single node with 8 NVIDIA H200 GPUs, while evaluations were performed using NVIDIA A100 GPUs. Each GRPO run trains for 50 epochs over 512 training problem instances with 8 rollouts per prompt; wall-clock training time ranges between 1–2 days. Reward computation runs on 16 CPU workers in parallel with the rollouts.

\section{PDE problems and solver bank}
\label{app:pde-bank}

This appendix specifies the eight PDE problems used throughout the paper and the selected solver bank (used during SFT phase) associated with each. The selected solvers are standard schemes drawn from numerical-methods literature, implemented in a common interface so that they can be enumerated and called uniformly. Exact verification of every solver implementation against a closed-form reference is not generally possible given the nonlinear and multidimensional problems in the bank, so we adopt the following verification protocol: self-convergence, the method of manufactured solutions (MMS) where an analytic forced solution is available, and cross-validation against published reference trajectories, which we view as adequate within our current scope and as a starting point in which future work can build on.

\subsection{PDE problems}
\label{app:pdes}

We consider eight PDEs spanning linear hyperbolic transport, nonlinear conservation laws, parabolic reaction--diffusion, diffusion--sorption with mixed boundary conditions, steady variable-coefficient elliptic problems, and the incompressible Navier--Stokes and shallow-water systems in two dimensions. \Cref{tab:pdes} summarizes the PDE type, geometry, boundary conditions, and expected solution output shapes for each. Throughout, subscripts denote partial derivatives, e.g., \(u_t=\partial u/\partial t\), \(u_x=\partial u/\partial x\), and \(u_{xx}=\partial^2 u/\partial x^2\), with the same convention applying to other scalar, vector, and conserved variables.

\paragraph{1D linear advection.}
We solve $u_t + \beta\,u_x = 0$ on $x \in [0,1)$ with periodic boundary conditions. We use \(\beta\in\{0.1,0.4,1.0,2.0\}\), uniform grids with \(N\in\{64,128\}\), output times
\(T\in\{51,101\}\), and final time \(t_{\mathrm{final}}=2.0\). Initial conditions are smooth periodic profiles sampled from four families: single sine, two-mode superposition, multimode waveform, and absolute-valued windowed waveform. The analytical solution is $u(x,t) = u_0(x - \beta t \bmod 1)$.

\paragraph{1D viscous Burgers.}
We solve $u_t + \partial_x\!\left(\frac{u^2}{2}\right) = \frac{\nu}{\pi}u_{xx}$ on $x \in [0,1)$ with periodic boundary conditions. Viscosity parameter is sampled from  $\nu \in \{10^{-3}, 10^{-2}, 10^{-1}, 1\}$. We use uniform grids with \(N\in\{64,128\}\), output times \(T\in\{51,101\}\), and \(t_{\mathrm{final}}=1.0\). 
Initial conditions are sampled from smooth periodic profiles: single-sine, two-mode, steeper-multimode families. As $\nu$ decreases, solutions develop steeper gradients, and the positive viscosity keeps the equation in the viscous regime within $t \in [0,1]$.

\paragraph{1D reaction--diffusion (Fisher--KPP).}
We solve $u_t = \nu\,u_{xx} + \rho\,u(1 - u)$ on $x \in [0,1)$ with periodic boundary conditions. Parameters are sampled from
\((\nu,\rho)\in\{0.5,1,2,5\}\times\{1,2,5,10\}\). We use uniform grids with \(N\in\{64,128\}\), output times \(T\in\{51,101\}\), and \(t_{\mathrm{final}}=1.0\). Initial conditions are sampled from single-sine, multimode, and front-like periodic families. This semilinear parabolic problem combines diffusion with logistic growth.

\paragraph{1D nonlinear diffusion--sorption.}
We solve $u_t = \dfrac{D}{R(u)}\,u_{xx}$ with retardation factor $R(u) = 1 + \frac{1 - \phi}{\phi}\,\rho_s\,k_f\,n_f\,(u + \varepsilon)^{n_f - 1}$, on $x \in [0,1]$ with the mixed boundary conditions $u(t,0) = u_{\mathrm{sol}}$ and $\partial_x u(t,1) = 0$. Parameters are sampled from
\(D\in\{10^{-4},5\times10^{-4},1.5\times10^{-3}\}\) and
\(k_f\in\{10^{-4},3.5\times10^{-4},10^{-3}\}\), with fixed
\(\phi=0.29\), \(\rho_s=2880\), \(n_f=0.874\), \(u_{\mathrm{sol}}=1\), and
\(\varepsilon=10^{-6}\). We use \(N\in\{64,128\}\), output times \(T\in\{51,101\}\), and \(t_{\mathrm{final}}=500.0\). Initial conditions are spatially uniform low-, mid-, and high-concentration
states. The PDEBench-inspired problem~\citep{takamoto2022pdebench} tests nonlinear diffusion with a concentration-dependent Freundlich retardation factor and mixed boundary conditions.

\paragraph{2D two-species reaction--diffusion.}
We solve $u_t = D_u\,\Delta u + (u - u^3 - k - v)$, $v_t = D_v\,\Delta v + (u - v)$, on $(x,y) \in [-1, 1]^2$ with homogeneous Neumann boundary conditions. Parameters are sampled from $(D_u, D_v, k) \in \{0.001, 0.002\} \times \{0.005\} \times \{0.005\}$. We use \(N_x,N_y\in\{32,48\}\), output times \(T\in\{31,51\}\), and \(t_{\mathrm{final}}=5.0\). Initial fields are sampled from Gaussian noise, blob pair, and striped noise families. The cubic activator and linear inhibitor produce labyrinthine and spotted patterns characteristic of FitzHugh--Nagumo-type activator--inhibitor dynamics~\citep{barkley1991model}.

\paragraph{2D Darcy flow.}
We solve the steady-state elliptic problem $-\nabla \cdot(a(x,y)\,\nabla u) = \beta$ on \((x,y)\in[0,1]^2\) with zero Dirichlet boundary conditions \(u=0\) on \(\partial\Omega\). The forcing parameter is sampled from \(\beta\in\{10^{-2},10^{-1},1,10,10^2\}\), the permeability field \(a(x,y)\) is sampled from a heterogeneous family: random blobs, checkerboard, and channelized, and we use
\(N_x,N_y\in\{32,48\}\). The solution output is expected to be the final steady-state field with shape \([B,N_x,N_y]\). This is the only steady-state problem in the PDE problems and solver bank.

\paragraph{2D incompressible Navier--Stokes.}
We solve $\nabla \cdot \mathbf{v} = 0$ and $\mathbf{v}_t + (\mathbf{v} \cdot \nabla)\mathbf{v} = -\nabla p + \nu\,\Delta \mathbf{v} + \mathbf{f}$ on $[0,1]^2$ with no-slip velocity boundary conditions $\mathbf{v} = 0$ on $\partial\Omega$. The viscosity is sampled from \(\nu\in\{0.005,0.01,0.02\}\). The initial velocity is generated from a random streamfunction with
\(\tau_{\mathbf{v}}=-3\), producing a divergence-reduced field. The two forcing components are independently sampled Gaussian random fields with \(\tau_{\mathbf{f}}=-1\). Velocity and forcing scales are sampled from \(\{0.1,0.15,0.2\}\) and \(\{0.25,0.4,0.55\}\), respectively. We use grids \((N_x,N_y)\in\{(16,16),(24,24)\}\), output times \(T\in\{11,21\}\), and \(t_{\mathrm{final}}=1.0\).

\paragraph{2D shallow water (dam break).}
We solve the shallow water system in conservative variables \((h,hu,hv)\): \(h_t + (hu)_x + (hv)_y = 0\), \((hu)_t + \left((hu)^2/h + \tfrac{1}{2}g h^2\right)_x + \left((hu)(hv)/h\right)_y = 0\), and \((hv)_t + \left((hu)(hv)/h\right)_x + \left((hv)^2/h + \tfrac{1}{2}g h^2\right)_y = 0\) on $(x,y) \in [-2.5, 2.5]^2$ with extrapolation-style boundary treatment, $g \in \{1, 2\}$. We use \(N_x,N_y\in\{32,48\}\), output times \(T\in\{31,51\}\), and \(t_{\mathrm{final}}=1.0\). Initial conditions are sampled from radial dam-break states with dam radii \(0.35\), \(0.50\), or \(0.65\).

\begin{table}[h]
\centering
\small
\caption{PDE problems used in this work. $B$ denotes the batch dimension of initial conditions, $T$ the number of output temporal snapshots, and $N$ (or $N_x \times N_y$) the spatial resolution. For multi-field systems, the last dimension indexes the returned variables.}
\label{tab:pdes}
\begin{tabular}{@{}lllll@{}}
\toprule
PDE & Type & Geometry & Output shape & \# solvers \\
\midrule
advection1d           & lin.\ hyperbolic       & 1D periodic    & $[B, T, N]$         & 8 \\
burgers1d             & nonlin.\ viscous cons. & 1D periodic    & $[B, T, N]$         & 18 \\
reaction\_diffusion1d & semilin.\ parabolic    & 1D periodic    & $[B, T, N]$         & 6 \\
diffusion\_sorption1d & nonlin.\ retarded par. & 1D mixed BC    & $[B, T, N]$         & 8 \\
reaction\_diffusion2d & 2-species parabolic    & 2D Neumann     & $[B, T, N_x, N_y, 2]$ & 6 \\
darcy2d               & steady elliptic        & 2D Dirichlet   & $[B, N_x, N_y]$     & 8 \\
incompressible\_ns2d  & incomp.\ NS w/ proj.   & 2D Dirichlet   & $[B, T, N_x, N_y, 2]$ & 10 \\
shallow\_water2d      & hyperbolic system      & 2D extrapol.   & $[B, T, N_x, N_y, 3]$ & 7 \\
\bottomrule
\end{tabular}
\end{table}

\subsection{Solver bank organization}
\label{app:bank-organization}

\Cref{tab:pdes} summarizes the selected solver bank used for training and evaluation. We annotate each solver here by its numerical method family, spatial discretization, time integrator, stability controls, and implementation checks before using them in the solver bank for SFT. 

The selected solvers in the bank are designed to cover a representative set of standard numerical idioms associated with each problem class rather than variants of a single template. For hyperbolic and conservation-law problems, the bank includes semi-Lagrangian characteristic methods, finite-volume Godunov, Rusanov, HLL/HLLC, Roe-type, MUSCL, Lax--Wendroff, MacCormack, and
Kurganov--Tadmor schemes, paired with standard explicit and strong-stability-preserving time integrators
\citep{godunov1959finite, van1979towards, kurganov2000new, leveque2002finite, toro2013riemann, gottlieb2001strong}. For parabolic and reaction--diffusion systems, the bank includes method-of-lines finite differences, Strang splitting with exact reaction updates when available, Fourier and DCT spectral diffusion, IMEX schemes, and implicit linearized or nonlinear solves \citep{strang1968construction, ascher1997implicit, trefethen2000spectral, kassam2005fourth}. 

For elliptic and incompressible-flow problems, the bank includes
cell-centered finite-volume discretizations, Jacobi and Gauss--Seidel variants, SOR, CG, preconditioned CG, ILU-preconditioned solves, and Chorin-style projection methods with finite-difference and spectral pressure solvers \citep{chorin1967numerical, strang1986introduction, virtanen2020scipy}. Spectral solvers use standard dealiasing where appropriate \citep{orszag1971elimination}.

For each PDE problem, detailed curated solver bank tables are listed in \Cref{tab:adv-bank,tab:burgers-bank,tab:rd1d-bank,tab:ds-bank,tab:rd2d-bank,tab:darcy-bank,tab:ns-bank,tab:sw-bank}. The reference column records either the implementation source (e.g., PDEBench-derived anchor schemes) or canonical method references for the corresponding discretization family.

\begin{table}[h]
\centering
\small
\caption{Solver bank for 1D linear advection. Each row gives the spatial scheme, time integrator, and reference source from which the implementation can be derived.}
\label{tab:adv-bank}
\resizebox{\linewidth}{!}{
\begin{tabular}{@{}llll@{}}
\toprule
codename & spatial scheme & time integrator & reference \\
\midrule
\texttt{semi\_lagrangian}              & periodic linear interp.            & characteristic backtrace with interp. & \citet{staniforth1991semi} \\
\texttt{pdebench\_fd2\_midpoint\_vanleer} & limited MUSCL, Rusanov flux  & RK2 midpoint                   & \citet{takamoto2022pdebench,van1979towards}           \\
\texttt{pdebench\_rusanov2}            & MUSCL (minmod), Rusanov flux       & RK2 midpoint                   & \citet{takamoto2022pdebench,leveque2002finite}               \\
\texttt{central4\_rk4}                 & 4th-order centered FD              & RK4                            & \citet{leveque2007finite}               \\
\texttt{muscl\_rusanov\_ssprk3}        & limited MUSCL, Rusanov flux       & SSPRK3                         & \citet{van1979towards,leveque2002finite,gottlieb2001strong}            \\
\texttt{spectral\_rk4}                 & Fourier pseudospectral             & RK4 with CFL substeps          & \citet{trefethen2000spectral}        \\
\texttt{fourier\_exact\_shift}         & FFT phase shift                    & exact per output interval      & \citet{trefethen2000spectral}        \\
\texttt{fourier\_spectral\_rk4}        & Fourier pseudospectral             & RK4 with CFL substeps          & \citet{trefethen2000spectral}        \\
\bottomrule
\end{tabular}}
\end{table}

\begin{table}[h]
\centering
\small
\caption{Solver bank for 1D viscous Burgers. Each row gives the spatial scheme, time integrator, and reference source from which the implementation can be derived. Most entries combine a hyperbolic flux with a centered finite-difference Laplacian for viscosity; spectral and IMEX methods instead treat diffusion in Fourier space or with Crank--Nicolson. Time integrators marked with $\dagger$ couple the explicit advection and the implicit diffusion through operator splitting within CFL substeps.}
\label{tab:burgers-bank}
\resizebox{\linewidth}{!}{
\begin{tabular}{@{}llll@{}}
\toprule
codename & spatial scheme & time integrator & reference \\
\midrule
\texttt{godunov\_\{euler,ssprk2,ssprk3\}}        & exact Godunov flux + centered Laplacian       & forward Euler / SSPRK2 / SSPRK3 & \citet{godunov1959finite,leveque2002finite,gottlieb2001strong} \\
\texttt{lax\_friedrichs\_\{euler,ssprk2,ssprk3\}} & local Lax--Friedrichs flux + centered Laplacian & forward Euler / SSPRK2 / SSPRK3 & \citet{leveque2002finite,gottlieb2001strong} \\
\texttt{rusanov\_\{euler,ssprk2,ssprk3\}}        & Rusanov flux + centered Laplacian              & forward Euler / SSPRK2 / SSPRK3 & \citet{rusanov1962calculation,leveque2002finite,gottlieb2001strong} \\
\texttt{muscl\_rusanov\_\{euler,rk2,ssprk2,ssprk3\}} & MUSCL (minmod), Rusanov flux + Laplacian   & forward Euler / RK2 / SSPRK2 / SSPRK3 & \citet{van1979towards,leveque2002finite,gottlieb2001strong} \\
\texttt{lax\_wendroff}                           & two-step Lax--Wendroff + Laplacian             & single step with CFL substeps   & \citet{lax1959systems,leveque2002finite} \\
\texttt{maccormack}                              & MacCormack predictor--corrector + Laplacian    & predictor--corrector            & \citet{maccormack2003effect,leveque2002finite} \\
\texttt{imex\_cn\_fixed}                         & Rusanov flux + periodic Laplacian              & explicit advection + CN diffusion$^\dagger$ & \citet{ascher1997implicit,hundsdorfer2003numerical} \\
\texttt{spectral\_rk4\_dealias}                  & Fourier pseudospectral, $2/3$ dealias          & RK4                             & \citet{orszag1971elimination,trefethen2000spectral} \\
\texttt{spectral\_if\_rk4\_dealias}              & dealiased nonlinear flux, exact diffusion      & integrating-factor RK4          & \citet{orszag1971elimination,trefethen2000spectral,kassam2005fourth} \\
\bottomrule
\end{tabular}}
\end{table}

\begin{table}[h]
\centering
\small
\caption{Solver bank for 1D reaction--diffusion. Each row gives the spatial scheme, time integrator, and reference source from which the implementation can be derived. The bank includes both second-order finite-difference and Fourier-spectral treatments of the periodic diffusion operator.}
\label{tab:rd1d-bank}
\resizebox{\linewidth}{!}{
\begin{tabular}{@{}llll@{}}
\toprule
codename & spatial scheme & time integrator & reference \\
\midrule
\texttt{fd\_explicit\_\{euler,rk2,rk4\}} & 2nd-order centered periodic FD Laplacian & forward Euler / RK2 / RK4 on full RHS & \citet{leveque2007finite,hundsdorfer2003numerical} \\
\texttt{split\_strang\_fd\_rk2\_pes}     & centered FD diffusion + pointwise logistic reaction & Strang splitting; exact/PES reaction half-steps + RK2 diffusion & \citet{strang1968construction,hundsdorfer2003numerical} \\
\texttt{spectral\_strang\_exact\_reaction} & Fourier spectral diffusion              & Strang splitting with exact logistic reaction and exact spectral diffusion & \citet{strang1968construction,trefethen2000spectral} \\
\texttt{spectral\_imex\_exact\_diffusion}  & Fourier spectral diffusion              & RK4 reaction + exact spectral diffusion & \citet{hundsdorfer2003numerical,trefethen2000spectral} \\
\bottomrule
\end{tabular}}
\end{table}

\begin{table}[h]
\centering
\small
\caption{Solver bank for 1D nonlinear diffusion--sorption. Each row gives the spatial scheme, time integrator or nonlinear treatment, and reference source from which the implementation can be derived. All entries use the mixed left-Dirichlet/right-Neumann boundary conditions treatment.}
\label{tab:ds-bank}
\resizebox{\linewidth}{!}{
\begin{tabular}{@{}llll@{}}
\toprule
codename & spatial scheme & time integrator / nonlinear treatment & reference \\
\midrule
\texttt{euler\_fixed}                    & lagged-mobility FD Laplacian             & forward Euler & \citet{leveque2007finite} \\
\texttt{heun\_fixed}                     & lagged-mobility FD Laplacian             & Heun / RK2 & \citet{hundsdorfer2003numerical} \\
\texttt{rk4\_fixed}                      & lagged-mobility FD Laplacian             & classical RK4 & \citet{hundsdorfer2003numerical} \\
\texttt{backward\_euler\_lagged}         & lagged-linearized mixed-BC tridiagonal system & backward Euler & \citet{hundsdorfer2003numerical} \\
\texttt{crank\_nicolson\_lagged}         & lagged-linearized mixed-BC tridiagonal system & Crank--Nicolson & \citet{hundsdorfer2003numerical} \\
\texttt{backward\_euler\_newton\_tridiag} & nonlinear mixed-BC tridiagonal Jacobian  & Newton backward Euler & \citet{hundsdorfer2003numerical} \\
\texttt{crank\_nicolson\_newton\_tridiag} & nonlinear mixed-BC tridiagonal Jacobian  & Newton Crank--Nicolson & \citet{hundsdorfer2003numerical} \\
\texttt{mol\_bdf\_solve\_ivp\_mixedbc}   & mixed-BC method of lines with tridiagonal sparsity & adaptive BDF via \texttt{solve\_ivp} & \citet{virtanen2020scipy} \\
\bottomrule
\end{tabular}}
\end{table}

\begin{table}[h]
\centering
\small
\caption{Solver bank for 2D two-species reaction--diffusion. Each row gives the spatial scheme, time integrator, and reference source from which the implementation can be derived. Finite-difference entries use edge padding for homogeneous Neumann boundary conditions, while DCT entries diagonalize the discrete Neumann Laplacian in a cosine basis.}
\label{tab:rd2d-bank}
\resizebox{\linewidth}{!}{
\begin{tabular}{@{}llll@{}}
\toprule
codename & spatial scheme & time integrator & reference \\
\midrule
\texttt{fd\_explicit\_\{euler,rk2,rk4\}} & 5-point FD Laplacian with Neumann edge padding & forward Euler / RK2 / RK4 on full RHS & \citet{leveque2007finite,hundsdorfer2003numerical} \\
\texttt{strang\_rk4\_reaction\_cn\_sparse} & sparse 5-point Neumann Laplacian & Strang splitting; RK4 reaction + CN diffusion & \citet{strang1968construction,hundsdorfer2003numerical} \\
\texttt{dct\_neumann\_strang\_split}     & DCT-diagonalized Neumann Laplacian        & Strang splitting; RK4 reaction + exact DCT diffusion & \citet{strang1968construction,trefethen2000spectral} \\
\texttt{dct\_neumann\_imex\_rk2}         & DCT-diagonalized Neumann Laplacian        & RK2 reaction + exact DCT diffusion & \citet{hundsdorfer2003numerical,trefethen2000spectral} \\
\bottomrule
\end{tabular}}
\end{table}

\begin{table}[h]
\centering
\small
\caption{Solver bank for 2D Darcy flow. Each row gives the spatial discretization, linear solver, and reference source from which the implementation can be derived. All entries solve a cell-centered finite-volume system with arithmetic face-averaged permeability and zero Dirichlet boundary values.}
\label{tab:darcy-bank}
\resizebox{\linewidth}{!}{
\begin{tabular}{@{}llll@{}}
\toprule
codename & spatial discretization & linear solver & reference \\
\midrule
\texttt{steady\_state\_cg\_fv\_arithmetic}                    & 5-point cell-centered FV, arithmetic faces & conjugate gradient & \citet{eymard2000finite,hestenes1952methods} \\
\texttt{steady\_state\_jacobi\_fv\_arithmetic}                & 5-point cell-centered FV, arithmetic faces & Jacobi relaxation & \citet{eymard2000finite,saad2003iterative} \\
\texttt{steady\_state\_weighted\_jacobi\_fv\_arithmetic}      & 5-point cell-centered FV, arithmetic faces & weighted Jacobi ($\omega=0.75$) & \citet{eymard2000finite,saad2003iterative} \\
\texttt{steady\_state\_redblack\_gauss\_seidel\_fv\_arithmetic} & 5-point cell-centered FV, arithmetic faces & red--black Gauss--Seidel & \citet{eymard2000finite,saad2003iterative} \\
\texttt{steady\_state\_sor\_fv\_arithmetic}                   & 5-point cell-centered FV, arithmetic faces & SOR ($\omega=1.7$) & \citet{eymard2000finite,saad2003iterative} \\
\texttt{steady\_state\_pcg\_jacobi\_fv\_arithmetic}           & 5-point cell-centered FV, arithmetic faces & Jacobi-preconditioned CG & \citet{eymard2000finite,hestenes1952methods} \\
\texttt{geometric\_multigrid\_vcycle\_fv}                     & 5-point cell-centered FV, arithmetic faces & Jacobi-preconditioned CG with sparse-direct fallback & \citet{eymard2000finite,hestenes1952methods,saad2003iterative,virtanen2020scipy} \\
\texttt{pcg\_ilu\_fv}                                         & 5-point cell-centered FV, arithmetic faces & ILU-preconditioned CG with sparse-direct fallback & \citet{eymard2000finite,hestenes1952methods,saad2003iterative,virtanen2020scipy} \\
\bottomrule
\end{tabular}}
\end{table}

\begin{table}[h]
\centering
\small
\caption{Solver bank for 2D incompressible Navier--Stokes. Each row gives the advection/diffusion discretization, projection and time integration, and reference source from which the implementation can be derived.}
\label{tab:ns-bank}
\resizebox{\linewidth}{!}{
\begin{tabular}{@{}llll@{}}
\toprule
codename & advection / diffusion discretization & projection and time integrator & reference \\
\midrule
\texttt{chorin\_centered\_euler}       & centered finite differences              & Chorin projection, forward Euler, Jacobi pressure solve & \citet{chorin1967numerical,ferziger2002computational} \\
\texttt{chorin\_centered\_rk2}         & centered finite differences              & Chorin projection, RK2, Jacobi pressure solve & \citet{chorin1967numerical,ferziger2002computational} \\
\texttt{chorin\_skewsym\_euler}        & skew-symmetric finite-difference advection & Chorin projection, forward Euler, Jacobi pressure solve & \citet{chorin1967numerical,ferziger2002computational} \\
\texttt{semiimplicit\_centered\_euler} & centered advection with semi-implicit diffusion & projection IMEX Euler, Jacobi pressure solve & \citet{chorin1967numerical,ascher1997implicit,ferziger2002computational} \\
\texttt{imex\_cnab2\_skewsym}          & skew-symmetric advection + CN diffusion  & CNAB2 projection, Jacobi pressure solve & \citet{ascher1997implicit,ferziger2002computational} \\
\texttt{dst\_projection\_cnab2\_runtime} & centered/skew-symmetric projection form & CNAB2 projection with DST Poisson/Helmholtz solves & \citet{chorin1967numerical,ascher1997implicit,strang1986introduction,virtanen2020scipy} \\
\texttt{imex\_cnab2\_skewsym\_spectral} & skew-symmetric advection + CN diffusion & CNAB2 projection with DST Poisson/Helmholtz solves & \citet{ascher1997implicit,strang1986introduction} \\
\texttt{semiimplicit\_centered\_spectral} & centered advection + semi-implicit diffusion & IMEX Euler projection with DST solves & \citet{ascher1997implicit,strang1986introduction} \\
\texttt{mac\_projection\_centered}     & centered advection + semi-implicit diffusion (DST projection alias) & IMEX Euler projection with DST Poisson/Helmholtz solves & \citet{chorin1967numerical,ascher1997implicit,strang1986introduction,virtanen2020scipy} \\
\texttt{mac\_projection\_rk2}          & skew-symmetric advection + CN diffusion (DST projection alias) & CNAB2 projection with DST Poisson/Helmholtz solves & \citet{ascher1997implicit,strang1986introduction,virtanen2020scipy} \\
\bottomrule
\end{tabular}}
\end{table}

\begin{table}[h]
\centering
\small
\caption{Solver bank for 2D shallow-water. Each row gives the flux, reconstruction and time integration strategy, and reference source from which the implementation can be derived.}
\label{tab:sw-bank}
\resizebox{\linewidth}{!}{
\begin{tabular}{@{}llll@{}}
\toprule
codename & flux & reconstruction \& time integrator & reference \\
\midrule
\texttt{fv\_muscl\_hll\_ssprk3}             & HLL                 & MUSCL (MC), SSPRK3 & \citet{toro2013riemann,van1979towards,gottlieb2001strong} \\
\texttt{fv\_muscl\_rusanov\_\{ssprk2,ssprk3\}}  & Rusanov         & MUSCL (MC), SSPRK2 / SSPRK3 & \citet{toro2013riemann,van1979towards,gottlieb2001strong} \\
\texttt{kt\_central\_upwind\_muscl\_ssprk3} & Kurganov--Tadmor    & MUSCL (MC), SSPRK3 & \citet{kurganov2000new,van1979towards,gottlieb2001strong} \\
\texttt{weno5\_rusanov\_ssprk3\_pos}        & Rusanov             & MUSCL (MC), SSPRK3 & \citet{toro2013riemann,van1979towards,gottlieb2001strong} \\
\texttt{roe\_entropy\_fix\_muscl\_ssprk3}   & Roe with entropy fix & MUSCL (MC), SSPRK3 & \citet{roe1981approximate,toro2013riemann,van1979towards,gottlieb2001strong} \\
\texttt{hllc\_muscl\_ssprk3\_pos}           & HLLC                & MUSCL (MC), SSPRK3 & \citet{toro2013riemann,van1979towards,gottlieb2001strong} \\
\bottomrule
\end{tabular}}
\end{table}

\subsection{Numerical methods verification protocol}
\label{app:verification}

We audit the code implementations used in the solver bank meant for SFT. Exact verification against a closed-form reference is not generally possible for the nonlinear and multidimensional problems in the solver bank, so we use three independent diagnostics, ordered from cheapest to most citable. We view the combined evidence as adequate within the scope of this paper while acknowledging that more comprehensive verification (e.g., adaptively-refined reference solutions or cross-comparison against an independent verified code base) is a natural extension.

\begin{enumerate}
\item \textbf{Self-convergence.} Standard refinement triples in $N$ at fixed output times, asserting that errors decrease monotonically and that observed orders match the formal order of the scheme. Where applicable, we additionally require that solvers within a scheme family agree on smooth initial data to within their formal order at fixed resolution. We present the quantitative results in \Cref{fig:convergence_solver_nbank}.

\item \textbf{Method of manufactured solutions (MMS).} For each PDE we construct a smooth analytic solution $u_e$, derive the corresponding source term $s_e$ symbolically, and run each solver on the forced PDE. This produces \emph{absolute} error against the exact $u_e$ at each grid level rather than a self-consistency ratio, and detects sign errors and stencil bugs that self-convergence cannot. For Burgers we use $u_e(x,t) = a + b\,\sin(2\pi(x - ct))$; for Darcy we use $u_e(x,y) = \sin(\pi x)\sin(\pi y)$ with permeability $a_e(x,y) = 1 + \tfrac{1}{2}\cos(2\pi x)\cos(2\pi y)$. We report representative MMS results for the PDE cases below.

\begin{figure}
    \centering
    \includegraphics[width=\linewidth]{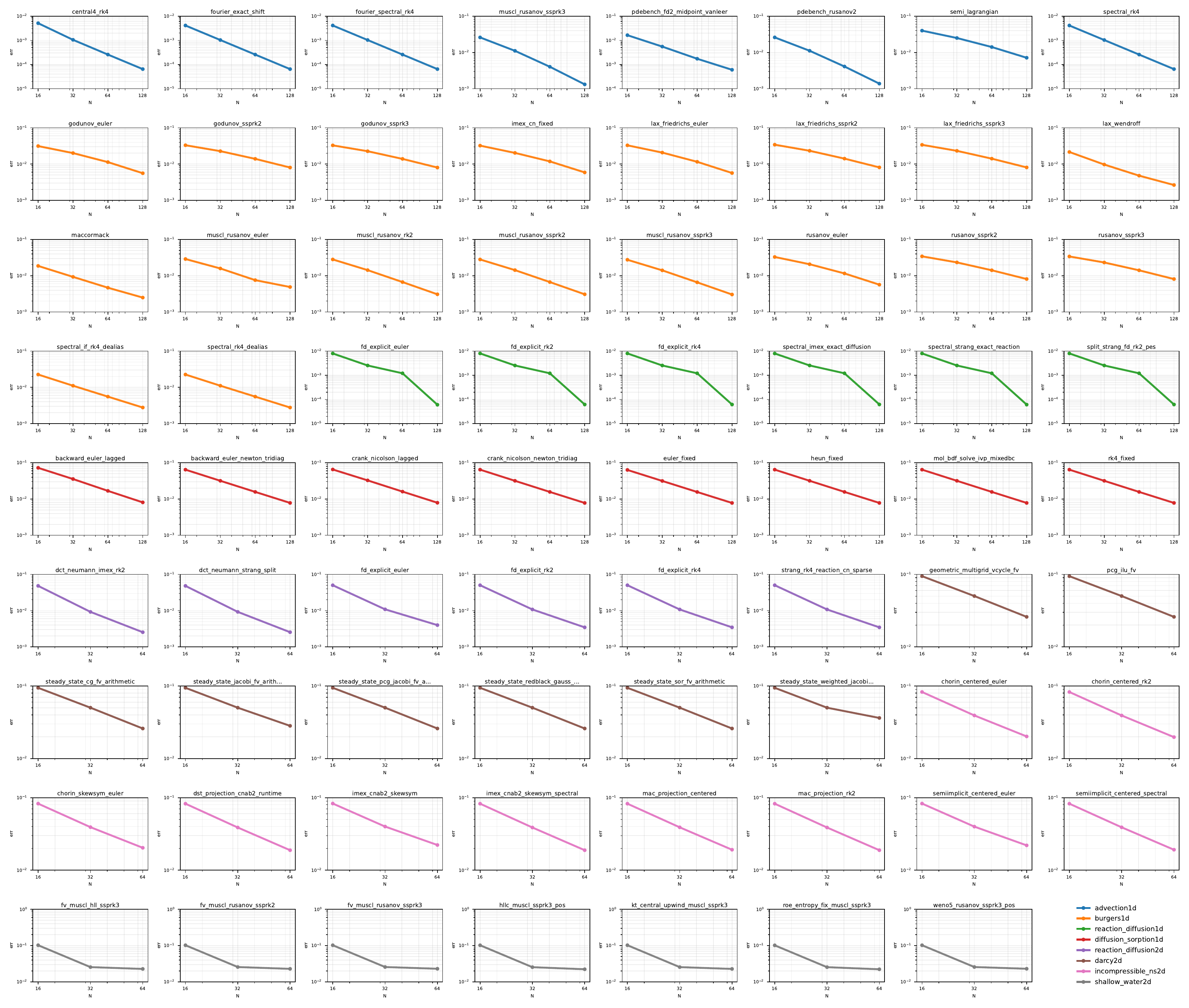}
    \caption{Self-convergence of the solver bank across the 8 PDE families. Each panel shows the relative pairwise error between consecutive grid resolutions as a function of the coarse resolution, using log-log axes. Lower error and a decreasing trend under refinement indicate numerically consistent solver behavior.}
    \label{fig:convergence_solver_nbank}
\end{figure}

\item \textbf{Cross-validation against published reference trajectories.}
For PDEs covered by PDEBench~\citep{takamoto2022pdebench}, we run the corresponding selected solvers on the published parameter settings and compare normalized RMSE against the released reference trajectories. This provides a check against an independent benchmark, complementing self-convergence and MMS.
\end{enumerate}

\Cref{tab:mms-burgers,tab:mms-darcy} report representative MMS results. Across the tested Burgers schemes, the observed refinement trends are consistent with the expected behavior of low-order, second-order, and spectral discretizations~\citep{leveque2002finite}. For Darcy flow, MMS results confirm that the cell-centered FV operator with arithmetic face averaging achieves observed order $2.00$ on smooth coefficients.


\begin{table}[h]
\centering
\small
\caption{MMS verification for representative 1D Burgers solvers, $\nu = 10^{-2}$, $t_{\mathrm{final}} = 0.2$, $\Delta t = 0.4\,\Delta x$. Observed orders are computed by linear regression of $\log \|u_h - u_e\|_{L^2}$ against $\log \Delta x$ across the refinement sequence.}
\label{tab:mms-burgers}
\resizebox{\linewidth}{!}{
\begin{tabular}{@{}lcccc@{}}
\toprule
solver & formal order & observed order & median $\|u - u_e\|_{L^2}$ at $N=256$ \\
\midrule
\texttt{lax\_friedrichs\_ssprk3}      & 1 & 0.99 & $7.6 \times 10^{-4}$ \\
\texttt{rusanov\_ssprk3}              & 1 & 1.00 & $7.6 \times 10^{-4}$ \\
\texttt{muscl\_rusanov\_ssprk3}       & 2 & 1.97 & $5.2 \times 10^{-5}$ \\
\texttt{lax\_wendroff}                & 2 & 1.99 & $4.8 \times 10^{-5}$ \\
\texttt{maccormack}                   & 2 & 1.99 & $4.7 \times 10^{-5}$ \\
\texttt{spectral\_if\_rk4\_dealias}   & superalgebraic & saturated & $7.7 \times 10^{-7}$ \\
\bottomrule
\end{tabular}}
\end{table}

\begin{table}[h]
\centering
\small
\caption{MMS verification for the 2D Darcy cell-centered FV discretization with arithmetic face averaging. Reference solve uses tight-tolerance preconditioned CG (relative residual $\le 10^{-12}$).}
\label{tab:mms-darcy}
\begin{tabular}{@{}cccc@{}}
\toprule
$N$ & $h$ & $\|u_h - u_e\|_{L^2}$ & observed order \\
\midrule
16  & $6.25 \times 10^{-2}$ & $1.12 \times 10^{-3}$ & --   \\
32  & $3.13 \times 10^{-2}$ & $2.80 \times 10^{-4}$ & 2.01 \\
64  & $1.56 \times 10^{-2}$ & $6.99 \times 10^{-5}$ & 2.00 \\
128 & $7.81 \times 10^{-3}$ & $1.75 \times 10^{-5}$ & 2.00 \\
\bottomrule
\end{tabular}
\end{table}

\FloatBarrier
\section{From Binary Verification to Expected Pass Rewards}
\label{app:expected_pass_reward}

The physical accuracy reward \(R_{\mathrm{traj}}\), instantiated through function-space trajectory error in \Cref{eq:traj_reward}, can also be viewed as a probabilistic analogue of binary RLVR. A standard binary verifier accepts a generated solver if its trajectory error is below a fixed tolerance,
\begin{equation}
    V_{\epsilon}(\widehat{u},u^\star)
    =
    \mathbf{1}\!\left\{
    \mathcal{L}_{\mathrm{traj}}(\widehat{u},u^\star)
    \le
    \epsilon
    \right\}.
\end{equation}
Instead of fixing a single tolerance, suppose the tolerance is drawn from a distribution
$\epsilon\sim\mathcal{D}$ with cumulative distribution function $F_{\mathcal{D}}$. The
corresponding expected binary reward is
\begin{equation}
    R_{\mathcal{D}}(\widehat{u},u^\star)
    =
    \mathbb{E}_{\epsilon\sim\mathcal{D}}
    \left[
    V_{\epsilon}(\widehat{u},u^\star)
    \right]
    =
    \mathbb{P}_{\epsilon\sim\mathcal{D}}
    \left(
    \epsilon
    \ge
    \mathcal{L}_{\mathrm{traj}}(\widehat{u},u^\star)
    \right).
\end{equation}
Thus, the dense reward induced by a stochastic tolerance is the survival function of the
tolerance distribution:
\begin{equation}
    R_{\mathcal{D}}(\widehat{u},u^\star)
    =
    1
    -
    F_{\mathcal{D}}
    \!\left(
    \mathcal{L}_{\mathrm{traj}}(\widehat{u},u^\star)
    \right).
    \label{eq:expected_binary_reward}
\end{equation}
This shows that a dense reward can be interpreted as the expected pass probability of a
binary verifier under uncertainty over the acceptance tolerance.

RLVP chooses an exponential tolerance distribution,
\begin{equation}
    \epsilon
    \sim
    \mathrm{Exp}(T_{\mathrm{traj}}),
    \qquad
    \mathbb{P}(\epsilon\ge t)
    =
    \exp(-t/T_{\mathrm{traj}})
    \quad
    \text{for } t\ge 0.
\end{equation}
Substituting this survival function into \eqref{eq:expected_binary_reward} gives
\begin{equation}
    \mathbb{E}_{\epsilon}
    \!\left[
    \mathbf{1}\!\left\{
    \mathcal{L}_{\mathrm{traj}}(\widehat{u},u^\star)
    \le
    \epsilon
    \right\}
    \right]
    =
    \exp\!\left(
    -\frac{
    \mathcal{L}_{\mathrm{traj}}(\widehat{u},u^\star)
    }{
    T_{\mathrm{traj}}
    }
    \right)
    =
    R_{\mathrm{traj}}(\widehat{u},u^\star).
    \label{eq:randomized_threshold}
\end{equation}
In this sense, $R_{\mathrm{traj}}$ is not merely a heuristic soft score. It is the exact
expected reward of a binary verifier whose acceptance tolerance is stochastic. The
exponential distribution is a minimally structured choice: it is the maximum-entropy
distribution on $\mathbb{R}_{\ge 0}$ with fixed mean and equivalently the waiting-time law
of a homogeneous Poisson process \citep{jaynes1957information, ross2014introduction}. It
places the highest density near strict tolerances while retaining an exponentially decaying
probability of more lenient thresholds. The temperature $T_{\mathrm{traj}}$ controls the
typical tolerance scale:
small $T_{\mathrm{traj}}$ concentrates the distribution near zero and makes the verifier
sharper, while large $T_{\mathrm{traj}}$ spreads probability over a wider range of
tolerances and gives smoother partial credit. In the limit
$T_{\mathrm{traj}}\to 0$, the reward approaches an exact-zero verifier,
$R_{\mathrm{traj}}\to\mathbf{1}\{\mathcal{L}_{\mathrm{traj}}=0\}$; in the limit
$T_{\mathrm{traj}}\to\infty$, $R_{\mathrm{traj}}\to 1$ for all finite errors. Other tolerance distributions would induce different dense rewards and we leave their analysis and inclusion as future work. 


The same Gibbs form appears in two classical contexts. In Markov chain Monte Carlo, the
Metropolis acceptance criterion accepts a proposal with energy increase $\Delta E$ with
probability $\exp(-\Delta E/T)$ \citep{metropolis1953equation}. In maximum-entropy
reinforcement learning, entropy-regularized optimal policies take an exponential form in
the value function, with the entropy temperature controlling the sharpness of the induced
distribution \citep{ziebart2008maximum, haarnoja2018soft}. Our use of the Gibbs form is
distinct from both: it is not a sampling acceptance rule and not a policy distribution, but
a verifier reward for generated PDE solvers. The derivation above makes the connection to
binary verification explicit: the Gibbs reward is the expected pass probability of a
binary verifier under an exponential distribution over tolerances.

The same stochastic-tolerance construction applies to any nonnegative diagnostic. For
physics residual consistency, RLVP uses the reference-relative residual loss in
\Cref{eq:phys_loss}; substituting this loss into the exponential expected-pass form yields
\Cref{eq:phys_reward}. Comparing to the reference residual absorbs discretization error
from the numerical reference solution and prevents the verifier from demanding a residual
smaller than the reference solver itself. A reference-free variant based directly on
$\rho(\widehat{u})$ recovers the canonical physics-residual regularization used in
physics-informed scientific machine learning
\citep{raissi2019physics, utkarsh2025physics,bastek2024physics}; we use the
reference-relative form for consistency with reference-solution evaluation.


\FloatBarrier
\section{Evaluation details}
\label{app:evaluation-details}
Let \(v_{p,c,s}\in\{0,1\}\) denote validity for prompt \(p\), case \(c\), and sample \(s\in\{1,\ldots,8\}\). 

We use the range-normalized evaluation error
\begin{equation}
    \mathrm{nRMSE}_{p,c,s}
    =
    \frac{
    \sqrt{\mathrm{mean}\left((\hat{u}_{p,c,s}-u^\star_{p,c})^2\right)}
    }{
    \max(u^\star_{p,c})-\min(u^\star_{p,c})
    },
\end{equation}
with the evaluator fallback to the RMS magnitude, and then to \(1\), when the reference range is numerically zero. A sample succeeds if it is valid and satisfies
\begin{equation}
    a_{p,c,s}
    =
    \mathbf{1}\left[
    v_{p,c,s}=1
    \;\wedge\;
    \mathrm{nRMSE}_{p,c,s}\leq 10^{-2}
    \right].
\end{equation}
We use a success threshold of nRMSE = \(10^{-2}\) for all PDEs in both seen and unseen tasks. 

We report combinatorial pass@\(k\). For each case, let \(q_{p,c}=\sum_{s=1}^{8}a_{p,c,s}\). With \(n=8\) samples,
\begin{equation}
    \widehat{\mathrm{pass@}k}_{p,c}
    =
    1-
    \frac{\binom{8-q_{p,c}}{k}}{\binom{8}{k}}.
\end{equation}
Thus pass@1 is \(q_{p,c}/8\), the expected success rate of one sampled solver, and pass@8 is \(\mathbf{1}[q_{p,c}>0]\), i.e. whether at least one of the 8 sampled solvers succeeds.

For continuous physical accuracy, we report median best-of-8 nRMSE. For samples that are invalid, missing, or non-finite, they receive a $1.0$ penalty with valid finite errors capped at $1.0$:

\begin{equation}
    \tilde{e}_{p,c,s}
    =
    \begin{cases}
    \min(\mathrm{nRMSE}_{p,c,s},1), & \text{if valid and finite},\\
    1, & \text{otherwise}.
    \end{cases}
\end{equation}

For each case, we take \(b^{@8}_{p,c}=\min_{s=1,\ldots,8}\tilde{e}_{p,c,s}\). Per-PDE nRMSE@8 is the median of \(b^{@8}_{p,c}\) over cases. Reported aggregate pass metrics are averaged over PDE prompts and aggregate nRMSE@8 is the median over PDE prompts (this prevents PDEs with more cases from dominating the aggregate).

\FloatBarrier
\section{Additional experimental details and figures}
\label{app:additional_results}

\begin{figure}[H]
    \centering
    \includegraphics[width=1\linewidth]{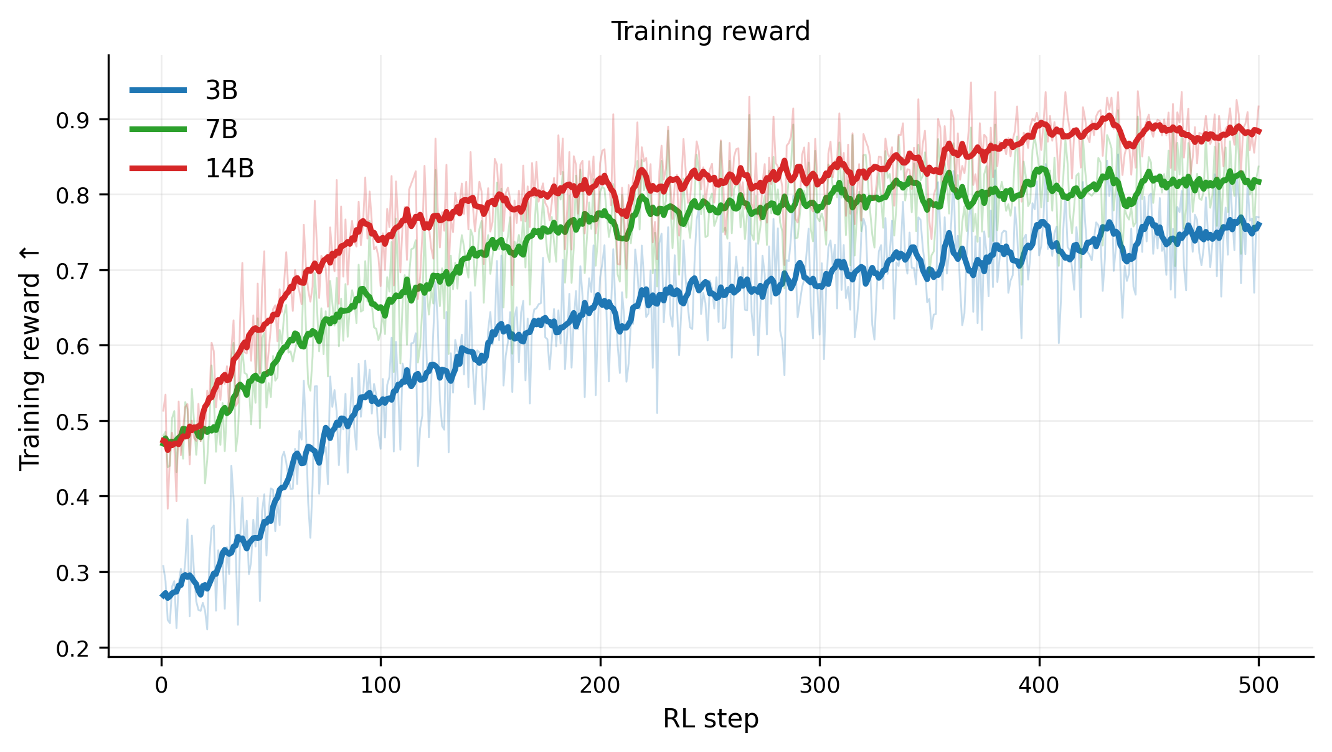}
    \caption{
    \textbf{Training reward over RL steps.}
    RLVP training reward increases smoothly across 3B, 7B, and 14B Qwen2.5-Coder-Instruct models. The curves show that verifier feedback provides a stable optimization signal after the SFT warm start.
    }
    \label{fig:training-reward-curves}
\end{figure}

\begin{figure}[t]
    \centering
    \includegraphics[width=1\linewidth]{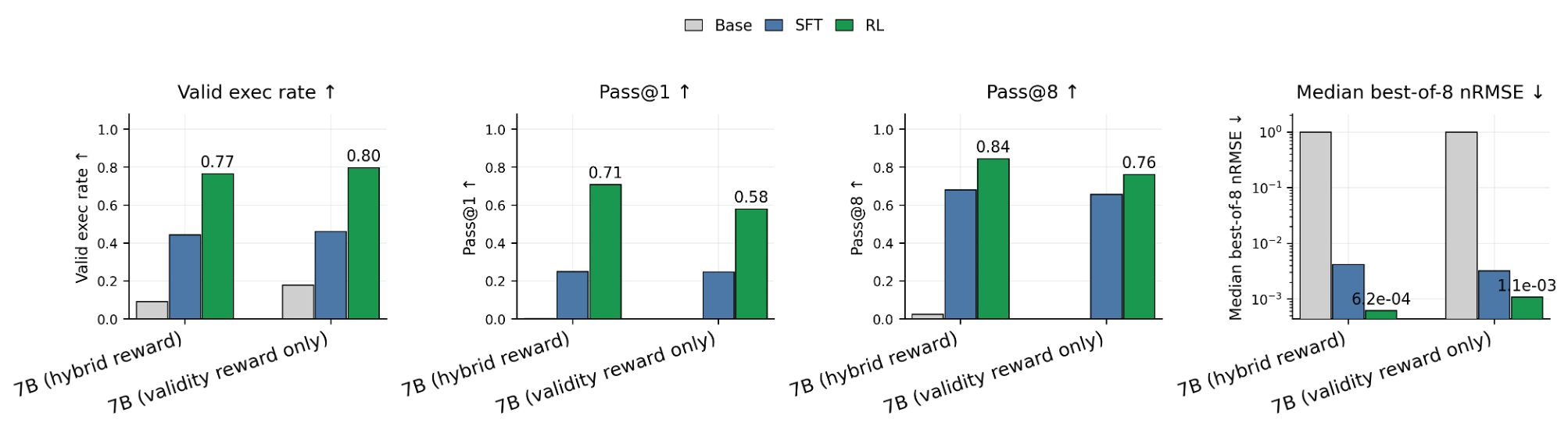}
    \caption{
    \textbf{Ablation of physical accuracy reward vs.\ validity-only reward for 8 seen PDE tasks.}
    We compare validity-only RL, \(R=V\), against validity plus physical accuracy, \(R=V R_{\mathrm{traj}}\), for the 7B model. Here, \(R_{\mathrm{traj}}\) is computed from function-space nRMSE against hidden references. Validity-only RL attains a similar valid execution rate, but adding \(R_{\mathrm{traj}}\) improves pass@1 (\(0.71\) vs.\ \(0.58\)), pass@8 (\(0.84\) vs.\ \(0.76\)), and median best-of-8 nRMSE (\(6.2\times10^{-4}\) vs.\ \(1.1\times10^{-3}\)). The physical accuracy reward improves solution quality beyond executable-code generation.
    }
    \label{fig:id_validity_nrmse_ablation}
\end{figure}

\begin{figure}[t]
    \centering
    \includegraphics[width=1\linewidth]{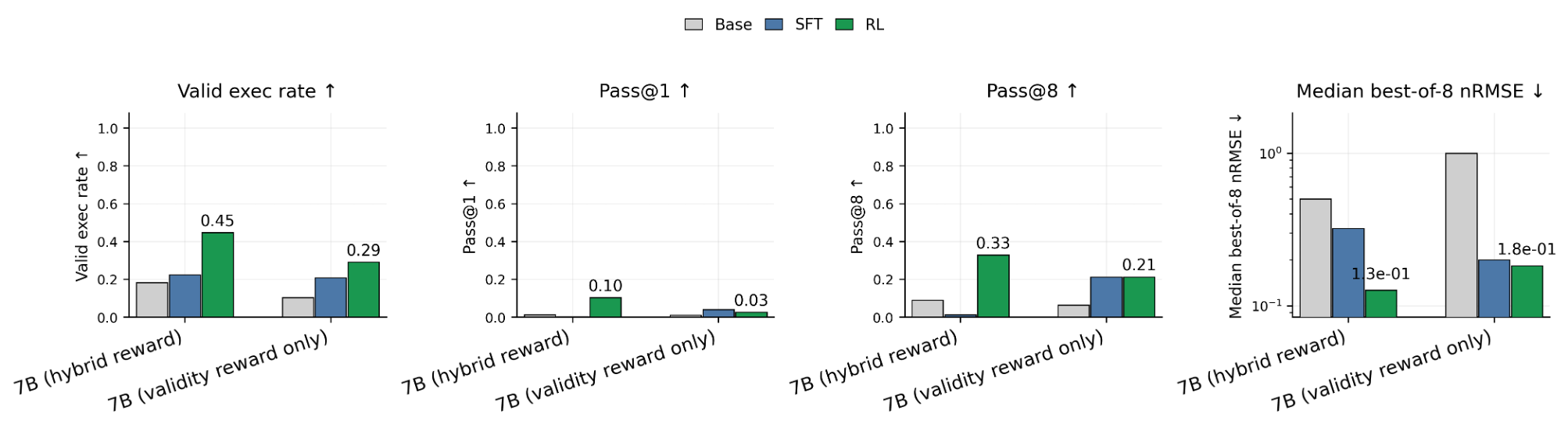}
    \caption{
    \textbf{Ablation of physical accuracy reward vs.\ validity-only reward for 10 unseen PDE tasks.}
    On the 10 held-out PDE prompts, adding \(R_{\mathrm{traj}}\) to validity improves valid execution rate (\(0.45\) vs.\ \(0.29\)), pass@1 (\(0.10\) vs.\ \(0.03\)), pass@8 (\(0.33\) vs.\ \(0.21\)), and median best-of-8 nRMSE (\(1.3\times10^{-1}\) vs.\ \(1.8\times10^{-1}\)) relative to validity-only RL. The physical accuracy reward is therefore important for cross-PDE transfer.
    }
    \label{fig:ood_validity_nrmse_ablation}
\end{figure}

\begin{figure}[H]
    \centering
    \includegraphics[width=0.8\linewidth]{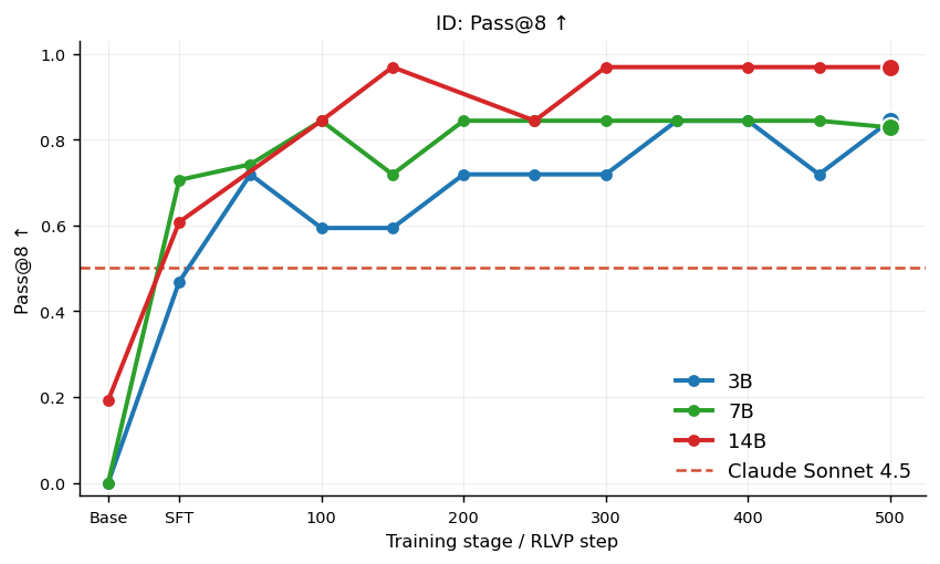}
    \caption{
    \textbf{Progress of pass@8 evaluation metrics during post-training.}
    Same setup as Figure~\ref{fig:id-progress}, but reporting combinatorial pass@8, comparing Base (Qwen2.5-Coder-Instruct), SFT, and RLVP post-trained checkpoints on the eight seen PDE tasks. Dashed line: Claude Sonnet 4.5 baseline evaluated with the same \(k=8\) sampling protocol.
    }
    \label{fig:id-pass8-app}
\end{figure}

\begin{figure}[t]
    \centering
    \includegraphics[width=0.9\linewidth]{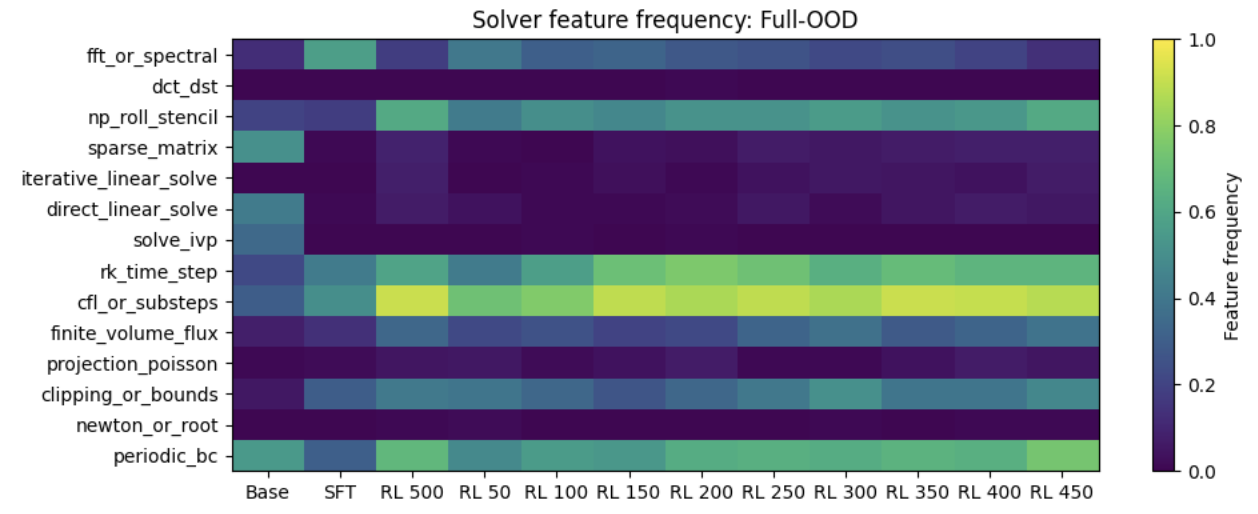}
    \caption{
    \textbf{Solver component frequencies across model checkpoints.}
    We parse generated solver code for numerical-method features. For solver generation on held-out PDE problems, RLVP post-trained checkpoints tend to more frequently use periodic boundary logic, \texttt{np.roll}-based stencils, CFL/substep control, and Runge--Kutta time stepping. These features align with the successful compositional transfer examples in Figure~\ref{fig:transfer-ard} and Appendix~\ref{app:transfer-examples}.
    }
    \label{fig:feature-frequency}
\end{figure}

\FloatBarrier
\section{Additional PDE solution and numerical method compositional transfer examples}
\label{app:transfer-examples}

We show code-level analysis examples for unseen PDE problems such as heat, 2D advection, and Cahn--Hilliard. For each held-out PDE, we select a representative solver generated by the RLVP post-trained LLM by success rate and median nRMSE across hidden PDE cases. We then compare it against the best SFT solver for the same held-out PDE and a corresponding matched seen PDE solver from RLVP. These examples are intended as possible qualitative evidences of selective numerical-scheme reuse and adaptation, rather than evidence that the post-trained model can solve arbitrary unseen PDEs.

\newpage
\begin{figure}[H]
    \centering
    \includegraphics[width=1.0\linewidth]{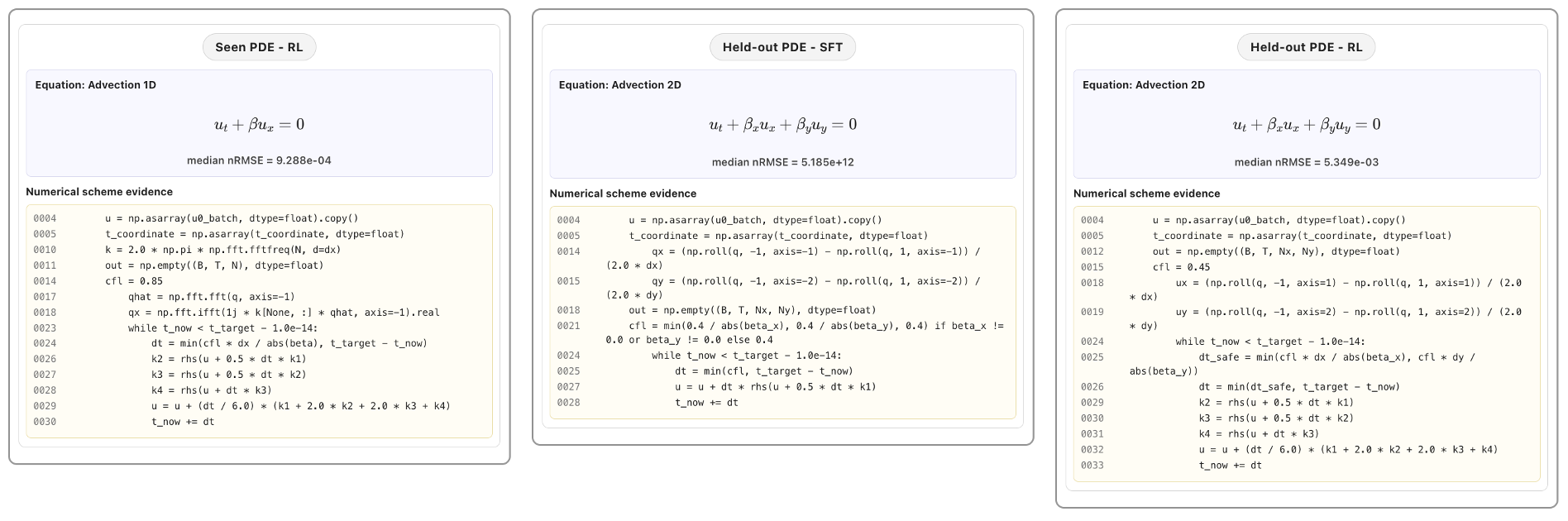}
    \caption{
    \textbf{Dimension lift transfer: 1D advection to 2D advection.}
    The matched seen problem gives the 1D advection solver generated by the RLVP post-trained model. The held-out RLVP solver lifts the transport structure to two dimensions using \(x/y\) derivative stencils, directional CFL control with \(dx\) and \(dy\), and RK4 time stepping, reaching median nRMSE \(5.349\times10^{-3}\). The SFT model generated solver is unstable, with median nRMSE \(5.185\times10^{12}\).
    }
    \label{fig:transfer-advection}
\end{figure}

\begin{figure}[H]
    \centering
    \includegraphics[width=1.0\linewidth]{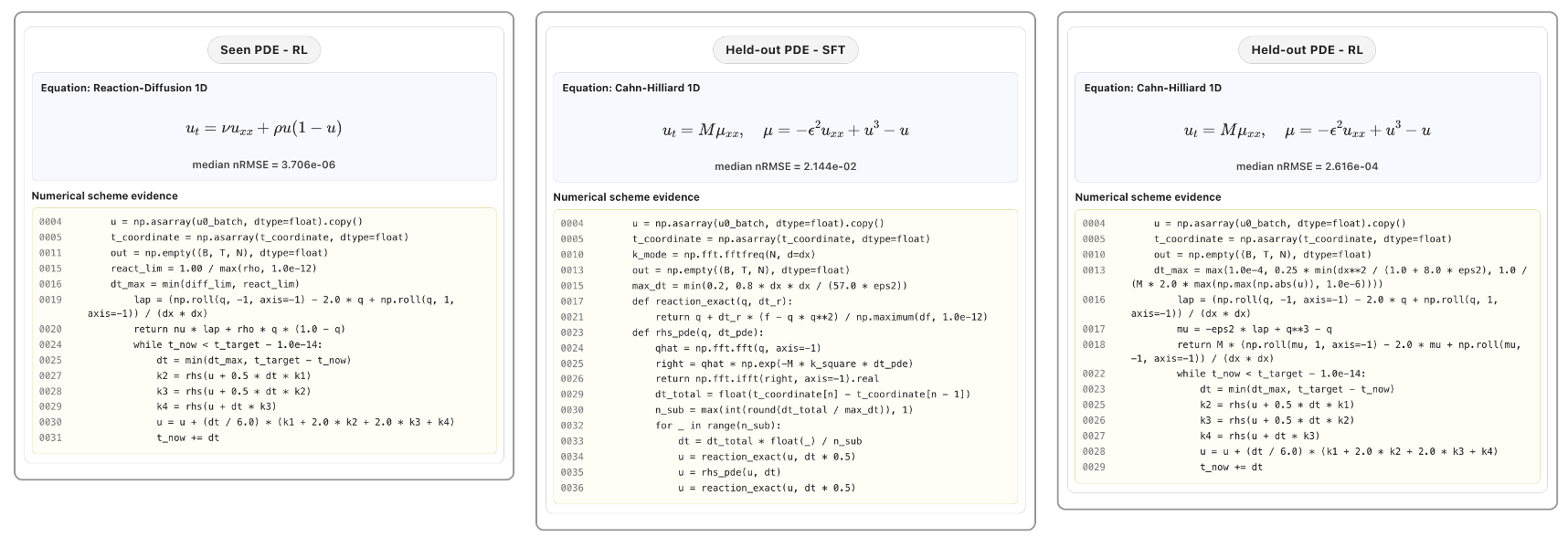}
    \caption{
    \textbf{Higher-order synthesis: Cahn--Hilliard.}
    The held-out Cahn--Hilliard PDE prompt requires a fourth-order structure. The solver generated by the RLVP post-trained model constructs the chemical potential \(\mu=-\epsilon^2 u_{xx}+u^3-u\) and applies a second Laplacian, reaching median nRMSE \(2.616\times10^{-4}\) compared with \(2.144\times10^{-2}\) for the SFT finetuned model. This suggests that the post-trained model could adapt familiar finite-difference primitives to assemble the higher-order operator requested by the prompt.
    }
    \label{fig:transfer-ch}
\end{figure}

\begin{figure}[H]
    \centering
    \includegraphics[width=1.0\linewidth]{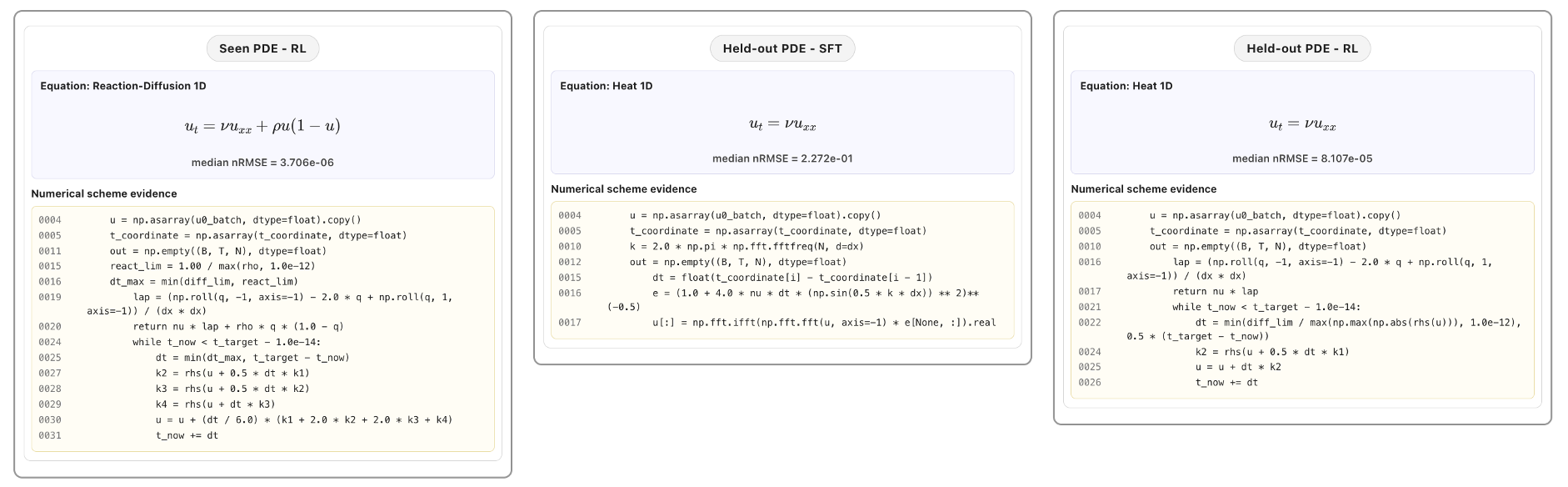}
    \caption{
    \textbf{Diffusion-subset transfer: reaction--diffusion to heat.}
    Heat is the diffusion-only subset of the seen reaction--diffusion family during SFT and RL. The solver generated by RLVP post-trained model drops the reaction source while retaining a local periodic Laplacian and explicit substepping, 
    improving median best-of-8 nRMSE from \(2.07\times10^{-1}\) for SFT to \(8.11\times10^{-5}\) after RL.
    }
    \label{fig:transfer-heat}
\end{figure}

\begin{table*}[t]
\centering
\scriptsize
\caption{
\textbf{Seen PDE problems per-PDE pass@8 and valid execution rate.}
This table complements Table~\ref{tab:id-main}. Pass@8 measures whether at least one of eight generated solvers succeeds on a case. Valid execution rate measures execution with correct shape and finite output. \textbf{Bold} marks the best meaningful value in each row and \underline{underline} marks the second-best meaningful value.
}
\label{tab:id-appendix}
\resizebox{\textwidth}{!}{%
\begin{tabular}{llcccccccccccc}
\toprule
 &  & \multicolumn{2}{c}{3B} & \multicolumn{2}{c}{7B} & \multicolumn{2}{c}{14B} & \multicolumn{1}{c}{Claude Sonnet 4.5} & \multicolumn{1}{c}{Qwen3-Coder-480B} & \multicolumn{1}{c}{DeepSeek V4 Flash} & \multicolumn{1}{c}{GPT-OSS-120B} & \multicolumn{1}{c}{Llama-3.1-405B} & \multicolumn{1}{c}{Llama-3.3-70B} \\
\cmidrule(lr){3-4} \cmidrule(lr){5-6} \cmidrule(lr){7-8} \cmidrule(lr){9-9} \cmidrule(lr){10-10} \cmidrule(lr){11-11} \cmidrule(lr){12-12} \cmidrule(lr){13-13} \cmidrule(lr){14-14}
Section & Metric & Base & RLVP & Base & RLVP & Base & RLVP & API & API & API & API & API & API \\
\midrule
ID per-PDE Pass@8 & advection1d & 0.00 & \textbf{1.00} & 0.00 & \underline{0.88} & 0.31 & \textbf{1.00} & 0.44 & 0.50 & 0.50 & 0.44 & 0.00 & 0.00 \\
 & burgers1d & 0.00 & \textbf{0.75} & 0.00 & \textbf{0.75} & 0.00 & \textbf{0.75} & \textbf{0.75} & \underline{0.62} & \textbf{0.75} & 0.50 & 0.50 & 0.00 \\
 & darcy2d & 0.00 & \textbf{1.00} & 0.00 & \textbf{1.00} & 0.00 & \textbf{1.00} & 0.00 & 0.00 & 0.00 & 0.00 & 0.00 & 0.00 \\
 & diffusion-sorption1d & 0.00 & \textbf{1.00} & 0.00 & \textbf{1.00} & \underline{0.22} & \textbf{1.00} & 0.00 & 0.00 & 0.00 & 0.00 & 0.00 & \underline{0.22} \\
 & incompressible-ns2d & 0.00 & 0.00 & 0.00 & 0.00 & 0.00 & \textbf{1.00} & \underline{0.33} & 0.00 & 0.00 & \textbf{1.00} & 0.00 & 0.00 \\
 & reaction-diffusion1d & 0.00 & \textbf{1.00} & 0.00 & \textbf{1.00} & \textbf{1.00} & \textbf{1.00} & \textbf{1.00} & \textbf{1.00} & \textbf{1.00} & 0.00 & \textbf{1.00} & \textbf{1.00} \\
 & reaction-diffusion2d & 0.00 & \textbf{1.00} & 0.00 & \textbf{1.00} & 0.00 & \textbf{1.00} & \textbf{1.00} & \textbf{1.00} & \textbf{1.00} & \textbf{1.00} & \underline{0.83} & 0.33 \\
 & shallow-water2d & 0.00 & \textbf{1.00} & 0.00 & \textbf{1.00} & 0.00 & \textbf{1.00} & \underline{0.50} & 0.00 & \underline{0.50} & \underline{0.50} & 0.00 & 0.00 \\
\midrule
ID per-PDE Valid exec rate & advection1d & 0.12 & \textbf{1.00} & 0.72 & \textbf{1.00} & 0.75 & 0.75 & \underline{0.97} & \textbf{1.00} & \textbf{1.00} & \textbf{1.00} & 0.53 & 0.41 \\
 & burgers1d & 0.00 & \textbf{1.00} & 0.12 & \textbf{1.00} & 0.37 & \textbf{1.00} & 0.52 & \underline{0.72} & 0.62 & 0.31 & 0.33 & 0.00 \\
 & darcy2d & 0.00 & \underline{0.75} & 0.00 & \underline{0.75} & 0.25 & \textbf{1.00} & \textbf{1.00} & 0.62 & \underline{0.75} & \textbf{1.00} & 0.25 & 0.38 \\
 & diffusion-sorption1d & 0.00 & \textbf{1.00} & 0.12 & 0.88 & 0.62 & \textbf{1.00} & \textbf{1.00} & 0.12 & \underline{0.96} & 0.88 & 0.06 & 0.12 \\
 & incompressible-ns2d & 0.00 & 0.00 & 0.00 & 0.25 & 0.00 & \textbf{0.62} & \underline{0.61} & \underline{0.61} & 0.38 & 0.12 & 0.00 & 0.00 \\
 & reaction-diffusion1d & 0.12 & 0.88 & 0.12 & \textbf{1.00} & 0.56 & \textbf{1.00} & 0.59 & 0.26 & \underline{0.96} & 0.10 & 0.36 & 0.25 \\
 & reaction-diffusion2d & 0.12 & 0.75 & 0.25 & \textbf{1.00} & 0.04 & \textbf{1.00} & \underline{0.88} & 0.38 & 0.62 & \textbf{1.00} & 0.50 & 0.50 \\
 & shallow-water2d & 0.12 & 0.12 & 0.00 & 0.50 & 0.00 & \textbf{1.00} & \underline{0.88} & 0.25 & 0.62 & 0.25 & 0.12 & 0.00 \\
\bottomrule
\end{tabular}%
}
\end{table*}


\begin{table*}[t]
\centering
\scriptsize
\caption{
\textbf{Aggregate performance on held-out PDE tasks.}
We report evaluation over 10 unseen PDE prompts. Best values within each model size are \textbf{bolded}.
}
\label{tab:ood-aggregate-app}
\resizebox{\textwidth}{!}{%
\begin{tabular}{lccccccccc}
\toprule
 & \multicolumn{3}{c}{3B} & \multicolumn{3}{c}{7B} & \multicolumn{3}{c}{14B} \\
\cmidrule(lr){2-4} \cmidrule(lr){5-7} \cmidrule(lr){8-10}
Metric & Base & SFT & RLVP & Base & SFT & RLVP & Base & SFT & RLVP \\
\midrule
Pass@1 $\uparrow$ & 0.01 & 0.03 & \textbf{0.07} & 0.01 & 0.04 & \textbf{0.11} & 0.11 & 0.14 & \textbf{0.21} \\
Pass@8 $\uparrow$ & 0.12 & 0.15 & \textbf{0.20} & 0.06 & 0.28 & \textbf{0.57} & 0.40 & \textbf{0.60} & 0.50 \\
Median-best nRMSE@8 $\downarrow$ & 3.02E-01 & 3.03E-01 & \textbf{2.64E-01} & 4.61E-01 & 1.45E-01 & \textbf{4.01E-03} & 3.72E-02 & \textbf{1.32E-03} & 1.49E-02 \\
\bottomrule
\end{tabular}%
}
\end{table*}

\begin{table*}[t]
\centering
\scriptsize
\caption{
\textbf{Unseen PDE performance per-PDE breakdown.}
This table expands Appendix Table~\ref{tab:ood-aggregate-app} with pass@1, pass@8, nRMSE@8, and valid execution rate for each held-out PDE problem. \textbf{Bold} marks the best value for each model size and \underline{underline} marks the second-best value.
}
\label{tab:ood-appendix}
\resizebox{\textwidth}{!}{%
\begin{tabular}{llccccccccc}
\toprule
 &  & \multicolumn{3}{c}{3B} & \multicolumn{3}{c}{7B} & \multicolumn{3}{c}{14B} \\
\cmidrule(lr){3-5} \cmidrule(lr){6-8} \cmidrule(lr){9-11}
Section & Metric & Base & SFT & RLVP & Base & SFT & RLVP & Base & SFT & RLVP \\
\midrule
Per-PDE Pass@1 & advection2d & 0.00 & 0.00 & 0.00 & 0.00 & 0.00 & \textbf{0.14} & 0.00 & \underline{0.12} & 0.00 \\
 & advection-diffusion1d & 0.00 & 0.00 & \textbf{0.01} & 0.00 & \textbf{0.04} & 0.00 & 0.00 & \textbf{0.35} & \underline{0.15} \\
 & advection-diffusion2d & 0.00 & 0.00 & 0.00 & 0.00 & \textbf{0.03} & 0.00 & \textbf{0.38} & \underline{0.12} & 0.00 \\
 & advection-reaction-diffusion1d & 0.00 & 0.00 & 0.00 & 0.00 & 0.00 & \textbf{0.25} & \textbf{0.38} & 0.00 & \underline{0.09} \\
 & allen-cahn2d & \textbf{0.03} & \underline{0.02} & 0.00 & 0.00 & \textbf{0.12} & \textbf{0.12} & 0.08 & \underline{0.25} & \textbf{0.38} \\
 & cahn-hilliard1d & 0.00 & \textbf{0.12} & 0.00 & 0.00 & \underline{0.03} & \textbf{0.30} & \textbf{0.21} & 0.00 & \underline{0.03} \\
 & darcy-reaction2d & 0.00 & 0.00 & 0.00 & 0.00 & 0.00 & 0.00 & 0.00 & 0.00 & 0.00 \\
 & gray-scott2d & \underline{0.10} & \textbf{0.12} & \textbf{0.12} & 0.00 & \textbf{0.12} & \textbf{0.12} & 0.08 & \underline{0.25} & \textbf{1.00} \\
 & heat1d & 0.00 & 0.00 & \textbf{0.59} & 0.00 & 0.00 & \textbf{0.17} & 0.00 & \underline{0.25} & \textbf{0.44} \\
 & kdv1d & \textbf{0.01} & 0.00 & 0.00 & \textbf{0.07} & 0.00 & 0.00 & 0.00 & 0.00 & 0.00 \\
\midrule
Per-PDE Pass@8 & advection2d & 0.00 & 0.00 & 0.00 & 0.00 & 0.00 & \textbf{0.75} & 0.00 & \textbf{1.00} & 0.00 \\
 & advection-diffusion1d & 0.00 & 0.00 & \textbf{0.04} & 0.00 & \textbf{0.33} & 0.00 & 0.00 & \textbf{1.00} & \textbf{1.00} \\
 & advection-diffusion2d & 0.00 & 0.00 & 0.00 & 0.00 & \textbf{0.25} & 0.00 & \textbf{1.00} & \textbf{1.00} & 0.00 \\
 & advection-reaction-diffusion1d & 0.00 & 0.00 & 0.00 & 0.00 & 0.00 & \textbf{1.00} & \textbf{1.00} & 0.00 & \underline{0.75} \\
 & allen-cahn2d & \textbf{0.25} & \underline{0.17} & 0.00 & 0.00 & \textbf{1.00} & \textbf{1.00} & \underline{0.33} & \textbf{1.00} & \textbf{1.00} \\
 & cahn-hilliard1d & 0.00 & \textbf{0.33} & 0.00 & 0.00 & \underline{0.25} & \textbf{1.00} & \textbf{1.00} & 0.00 & \underline{0.25} \\
 & darcy-reaction2d & 0.00 & 0.00 & 0.00 & 0.00 & 0.00 & 0.00 & 0.00 & 0.00 & 0.00 \\
 & gray-scott2d & \underline{0.83} & \textbf{1.00} & \textbf{1.00} & 0.00 & \textbf{1.00} & \textbf{1.00} & \underline{0.67} & \textbf{1.00} & \textbf{1.00} \\
 & heat1d & 0.00 & 0.00 & \textbf{1.00} & 0.00 & 0.00 & \textbf{1.00} & 0.00 & \textbf{1.00} & \textbf{1.00} \\
 & kdv1d & \textbf{0.11} & 0.00 & 0.00 & \textbf{0.56} & 0.00 & 0.00 & 0.00 & 0.00 & 0.00 \\
\midrule
Per-PDE nRMSE@8 & advection2d & 3.15e-01 & \underline{3.01e-01} & \textbf{2.63e-01} & \underline{2.69e-01} & 1.00e+00 & \textbf{5.35e-03} & \underline{2.35e-01} & \textbf{1.14e-14} & 2.91e-01 \\
 & advection-diffusion1d & 9.26e-01 & \underline{3.05e-01} & \textbf{9.01e-02} & 3.29e-01 & \textbf{5.07e-02} & \underline{2.20e-01} & \underline{1.50e-01} & \textbf{5.36e-04} & \textbf{5.36e-04} \\
 & advection-diffusion2d & \underline{2.89e-01} & \textbf{1.33e-01} & 1.00e+00 & 1.00e+00 & \textbf{1.34e-01} & 1.00e+00 & \textbf{1.63e-03} & \textbf{1.63e-03} & \underline{1.22e-01} \\
 & advection-reaction-diffusion1d & 6.46e-01 & \underline{4.59e-01} & \textbf{2.65e-01} & 5.71e-01 & \underline{5.25e-01} & \textbf{1.50e-03} & \textbf{9.95e-04} & 1.21e-01 & \underline{3.90e-03} \\
 & allen-cahn2d & \textbf{1.59e-02} & \underline{1.08e-01} & 2.43e-01 & 3.52e-01 & \textbf{3.44e-04} & \textbf{3.44e-04} & \underline{1.70e-02} & 3.44e-04 & \textbf{2.65e-04} \\
 & cahn-hilliard1d & \underline{3.58e-02} & \textbf{1.81e-02} & 1.00e+00 & 4.17e-02 & \underline{2.14e-02} & \textbf{2.62e-04} & \textbf{2.82e-04} & \underline{2.61e-02} & 1.00e+00 \\
 & darcy-reaction2d & \textbf{6.71e-01} & 1.00e+00 & 1.00e+00 & 1.00e+00 & \textbf{2.00e-01} & \underline{2.37e-01} & 1.00e+00 & 1.00e+00 & \textbf{2.59e-02} \\
 & gray-scott2d & 4.54e-03 & \textbf{1.13e-03} & \textbf{1.13e-03} & 1.00e+00 & \textbf{1.14e-03} & \underline{2.68e-03} & 4.50e-03 & \textbf{1.01e-03} & \textbf{1.01e-03} \\
 & heat1d & 1.00e+00 & \underline{3.13e-01} & \textbf{7.39e-05} & 1.00e+00 & \underline{2.07e-01} & \textbf{8.11e-05} & 1.00e+00 & \textbf{1.68e-06} & \underline{7.39e-05} \\
 & kdv1d & \textbf{9.49e-02} & 1.00e+00 & 1.00e+00 & \textbf{7.50e-03} & \underline{1.55e-01} & 1.00e+00 & \underline{5.75e-02} & \textbf{4.82e-02} & 1.00e+00 \\
\midrule
Per-PDE valid executable rate & advection2d & \textbf{0.25} & \textbf{0.25} & \underline{0.12} & \underline{0.19} & 0.12 & \textbf{0.88} & \textbf{0.50} & \underline{0.41} & 0.28 \\
 & advection-diffusion1d & 0.12 & \underline{0.16} & \textbf{0.31} & 0.25 & \textbf{0.75} & \underline{0.29} & 0.38 & \textbf{0.75} & \underline{0.72} \\
 & advection-diffusion2d & \underline{0.38} & \textbf{0.50} & 0.00 & 0.00 & \textbf{0.34} & 0.00 & \underline{0.47} & 0.25 & \textbf{0.69} \\
 & advection-reaction-diffusion1d & \underline{0.19} & 0.12 & 0.12 & \underline{0.25} & \underline{0.25} & \textbf{0.59} & 0.54 & \underline{0.75} & \textbf{0.82} \\
 & allen-cahn2d & \textbf{0.38} & \underline{0.25} & 0.12 & \textbf{0.25} & \underline{0.24} & 0.12 & 0.25 & \underline{0.75} & \textbf{0.88} \\
 & cahn-hilliard1d & \underline{0.12} & \textbf{0.62} & 0.00 & 0.12 & \underline{0.36} & \textbf{0.69} & \textbf{0.88} & \underline{0.41} & 0.06 \\
 & darcy-reaction2d & \textbf{0.12} & 0.00 & 0.00 & 0.00 & \underline{0.25} & \textbf{0.38} & 0.12 & 0.02 & \textbf{0.25} \\
 & gray-scott2d & \underline{0.12} & \textbf{0.25} & \underline{0.12} & 0.00 & \underline{0.25} & \textbf{0.38} & \underline{0.38} & 0.25 & \textbf{1.00} \\
 & heat1d & \underline{0.32} & 0.25 & \textbf{0.75} & 0.00 & \textbf{0.75} & \underline{0.50} & 0.00 & \textbf{0.71} & \underline{0.50} \\
 & kdv1d & \textbf{0.11} & 0.00 & 0.00 & \textbf{0.25} & \underline{0.12} & 0.04 & 0.14 & \textbf{0.62} & \underline{0.42} \\
\bottomrule
\end{tabular}%
}
\end{table*}

\FloatBarrier
\section{Example Solver Programs Generated by the Post-Trained Policy}
\label{app:generated_codes}

In this section, we show example snippets of generated solver codes for evaluated PDE prompts on the RLVP post-trained 7B model.

\begin{figure}[H]
    \centering
    \includegraphics[width=1\linewidth]{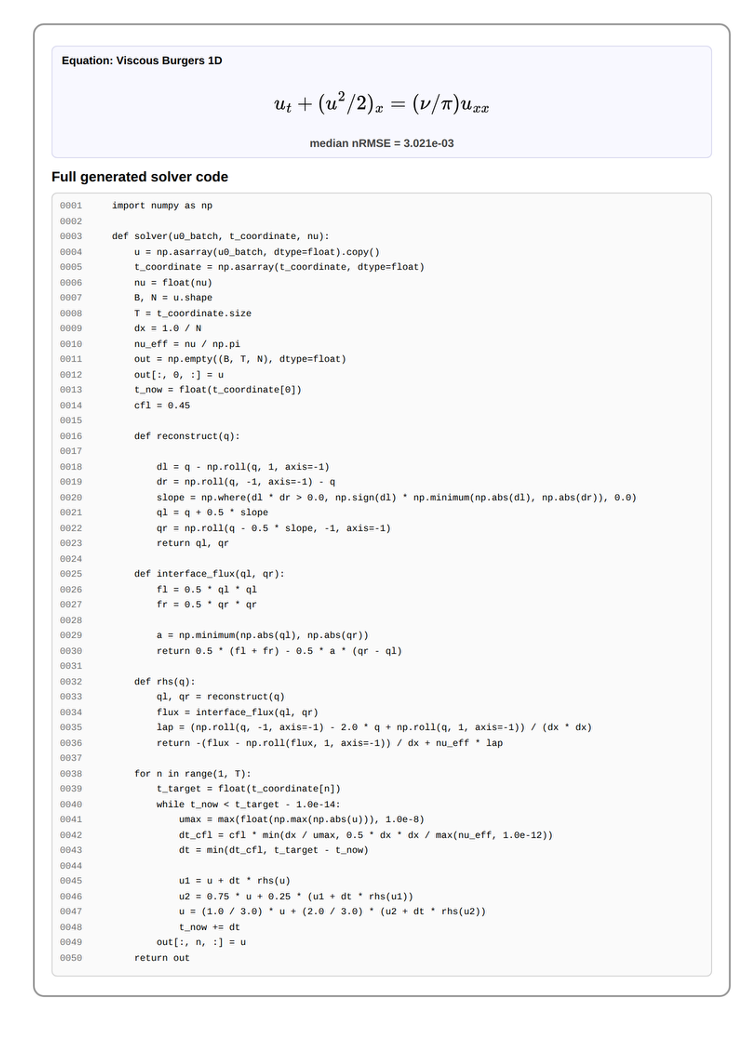}
    \caption{
    \textbf{Generated solver from the RLVP post-trained model for 1D viscous Burgers.}
    Best selected solver from the 7B RLVP checkpoint on the seen viscous Burgers task.
    }
    \label{fig:id-burgers1d-solver}
\end{figure}

\begin{figure}[H]
    \centering
    \includegraphics[width=1.0\linewidth]{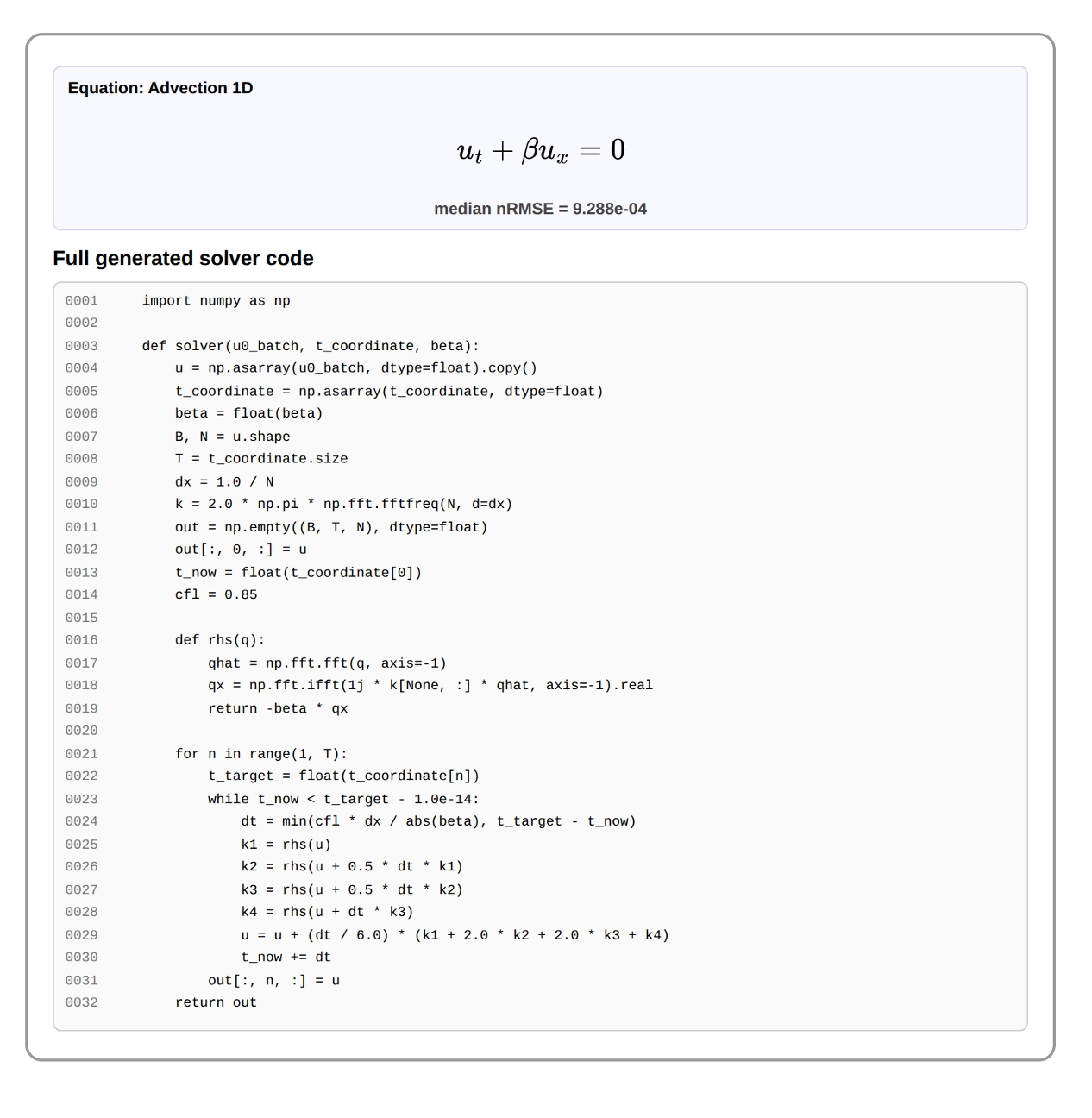}
    \caption{
    \textbf{Generated solver from the RLVP post-trained model for 1D advection.}
    Best selected solver from the 7B RLVP checkpoint on the seen 1D advection task.
    }
    \label{fig:id-advection1d-solver}
\end{figure}

\begin{figure}[H]
    \centering
    \includegraphics[width=1.0\linewidth]{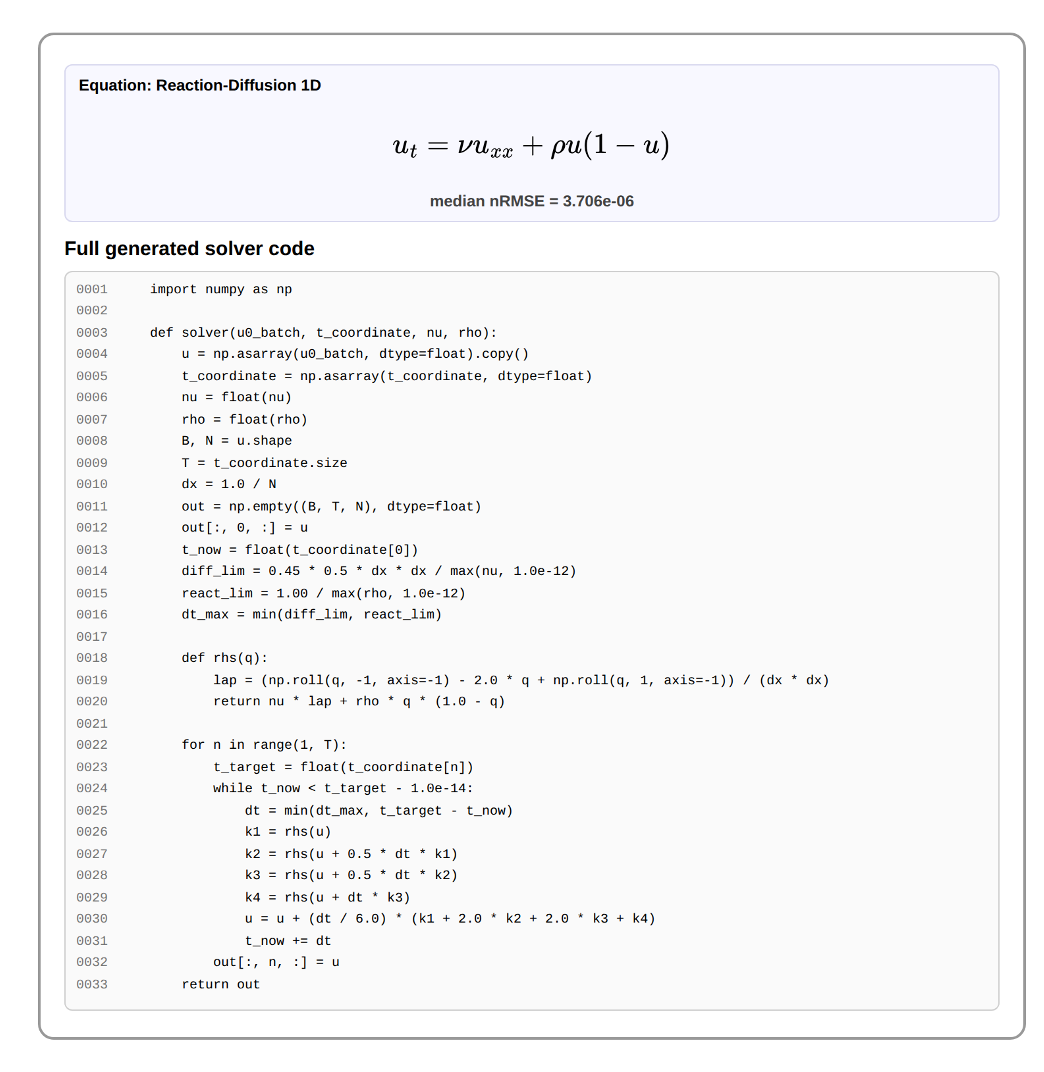}
    \caption{
    \textbf{Generated solver from the RLVP post-trained model for 1D reaction--diffusion.}
    Best selected solver from the 7B RLVP checkpoint on the seen reaction--diffusion task.
    }
    \label{fig:id-rd1d-solver}
\end{figure}

\begin{figure}[H]
    \centering
    \includegraphics[width=1.0\linewidth]{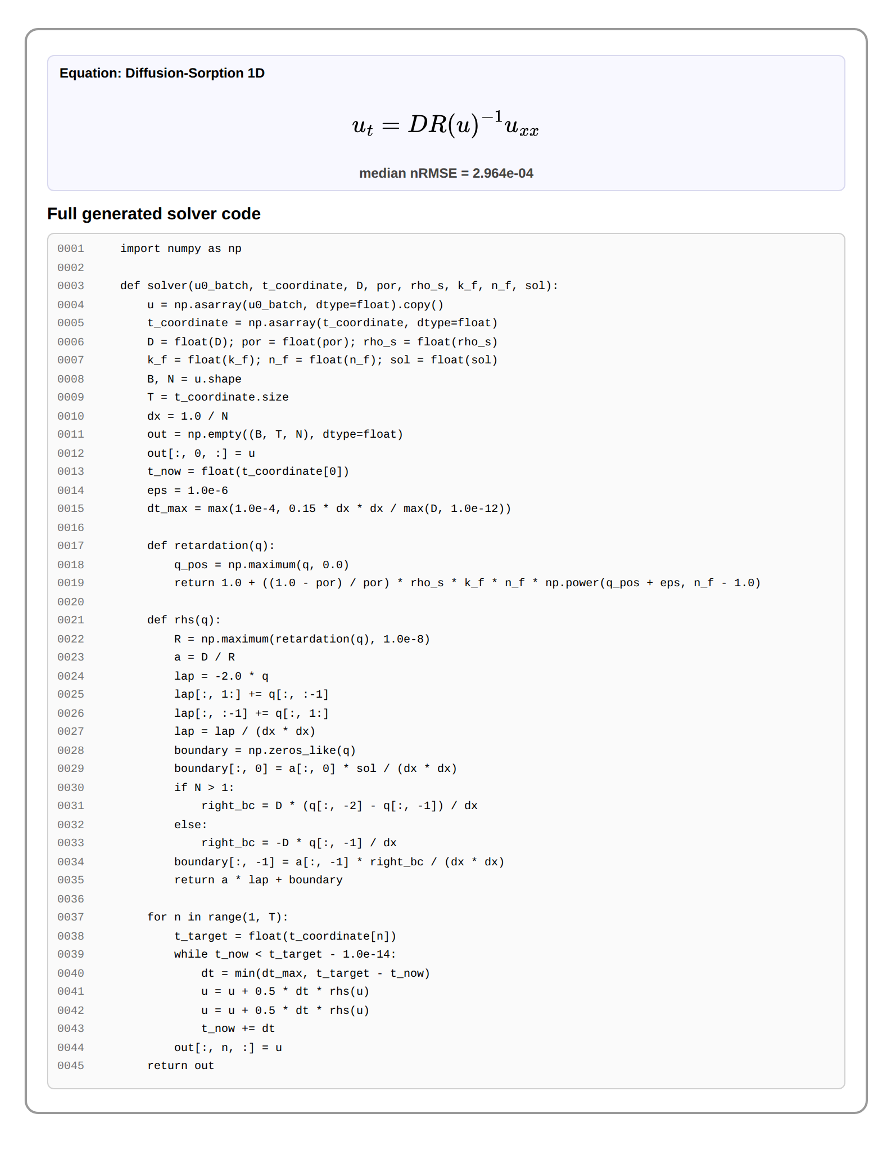}
    \caption{
    \textbf{Generated solver from the RLVP post-trained model for 1D diffusion--sorption.}
    Best selected solver from the 7B RLVP checkpoint on the seen diffusion--sorption task.
    }
    \label{fig:id-ds1d-solver}
\end{figure}

\begin{figure}[H]
    \centering
    \includegraphics[width=1\linewidth]{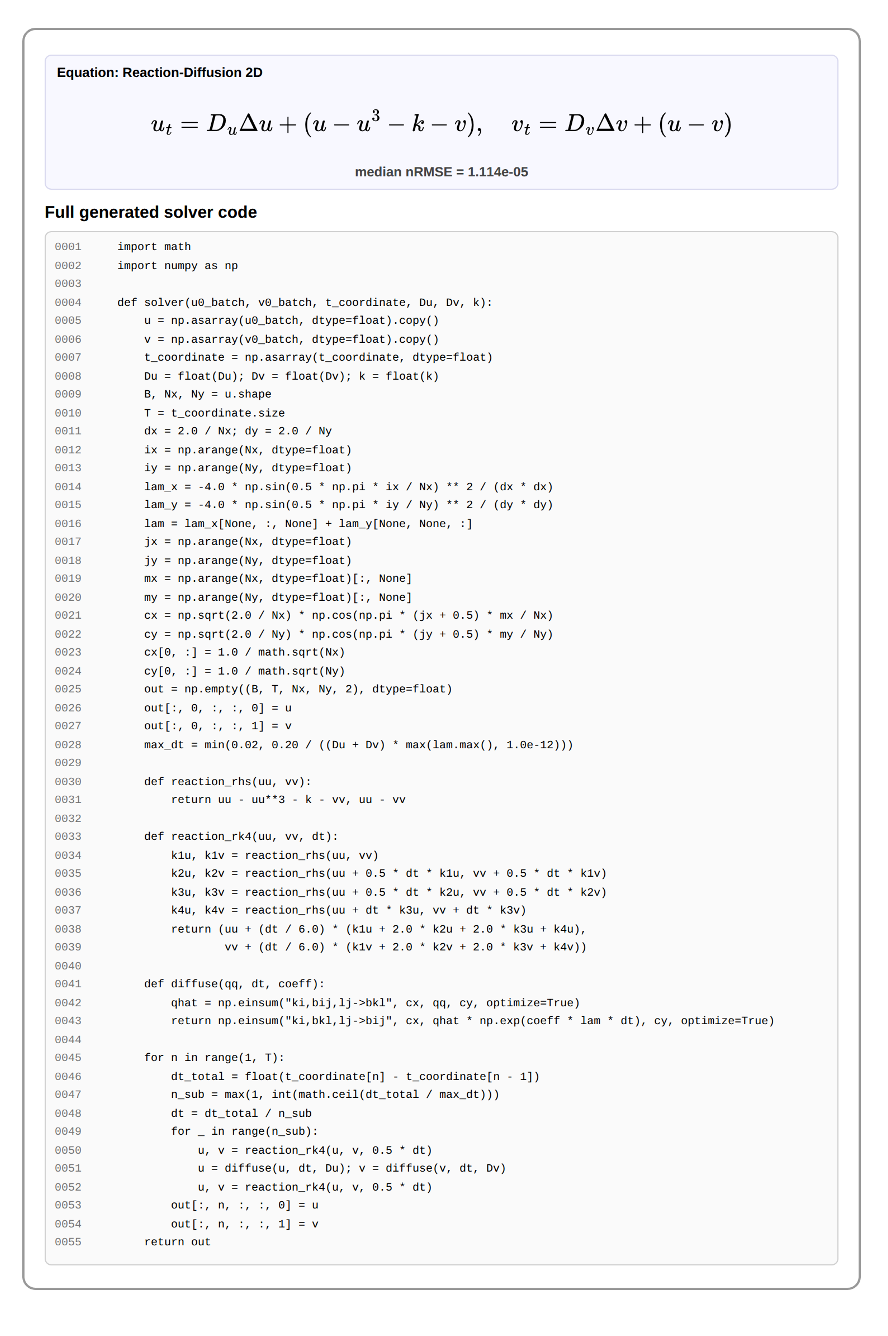}
    \caption{
    \textbf{Generated solver from the RLVP post-trained model for 2D reaction--diffusion.}
    Best selected solver from the 7B RLVP checkpoint on the seen 2D reaction--diffusion task.
    }
    \label{fig:id-rd2d-solver}
\end{figure}

\begin{figure}[H]
    \centering
    \includegraphics[width=1.0\linewidth]{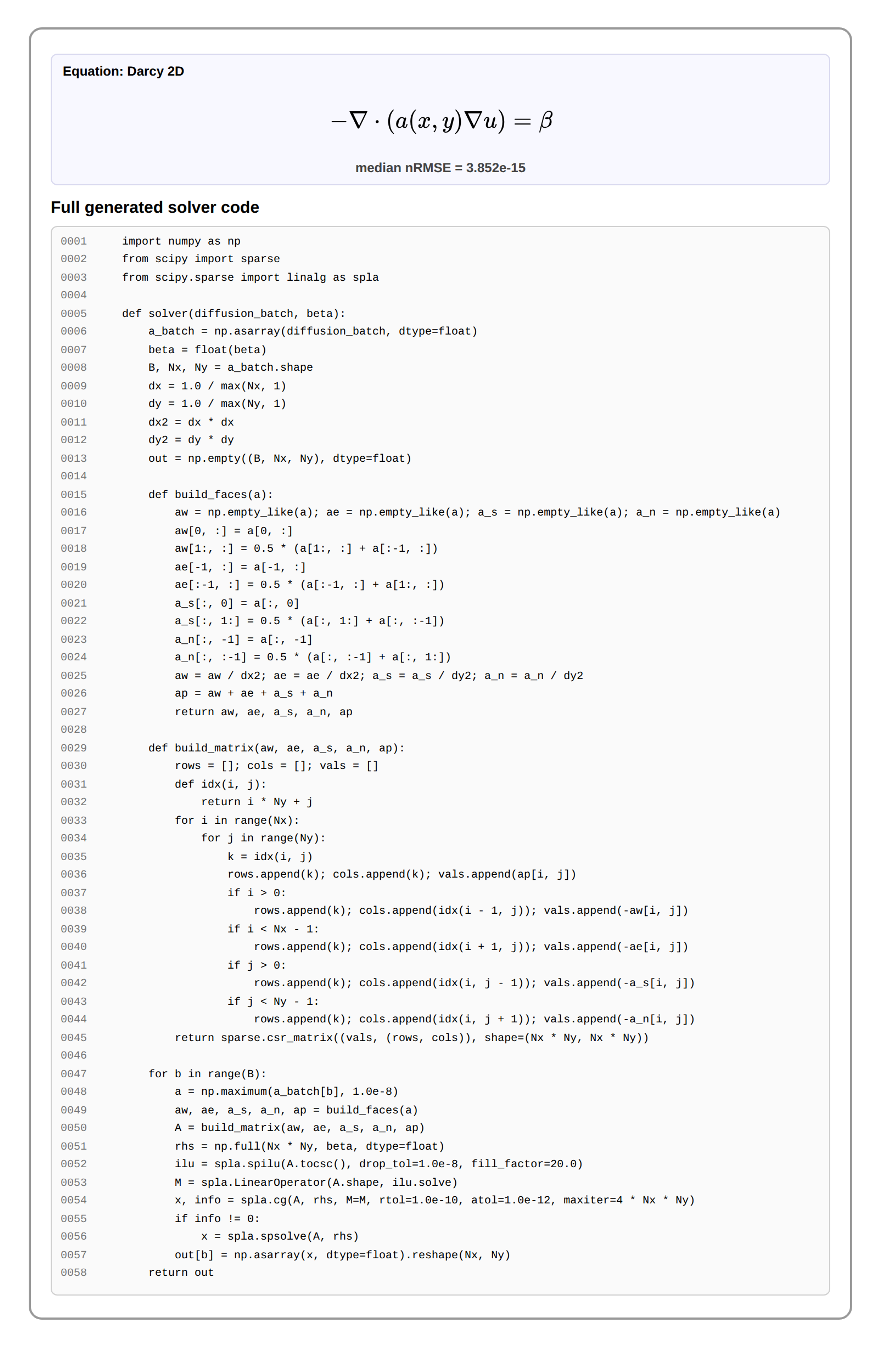}
    \caption{
    \textbf{Generated solver from the RLVP post-trained model for 2D Darcy flow.}
    Best selected solver from the 7B RLVP checkpoint on the seen Darcy flow task.
    }
    \label{fig:id-darcy2d-solver}
\end{figure}

\begin{figure}[H]
    \centering
    \includegraphics[width=0.65\linewidth]{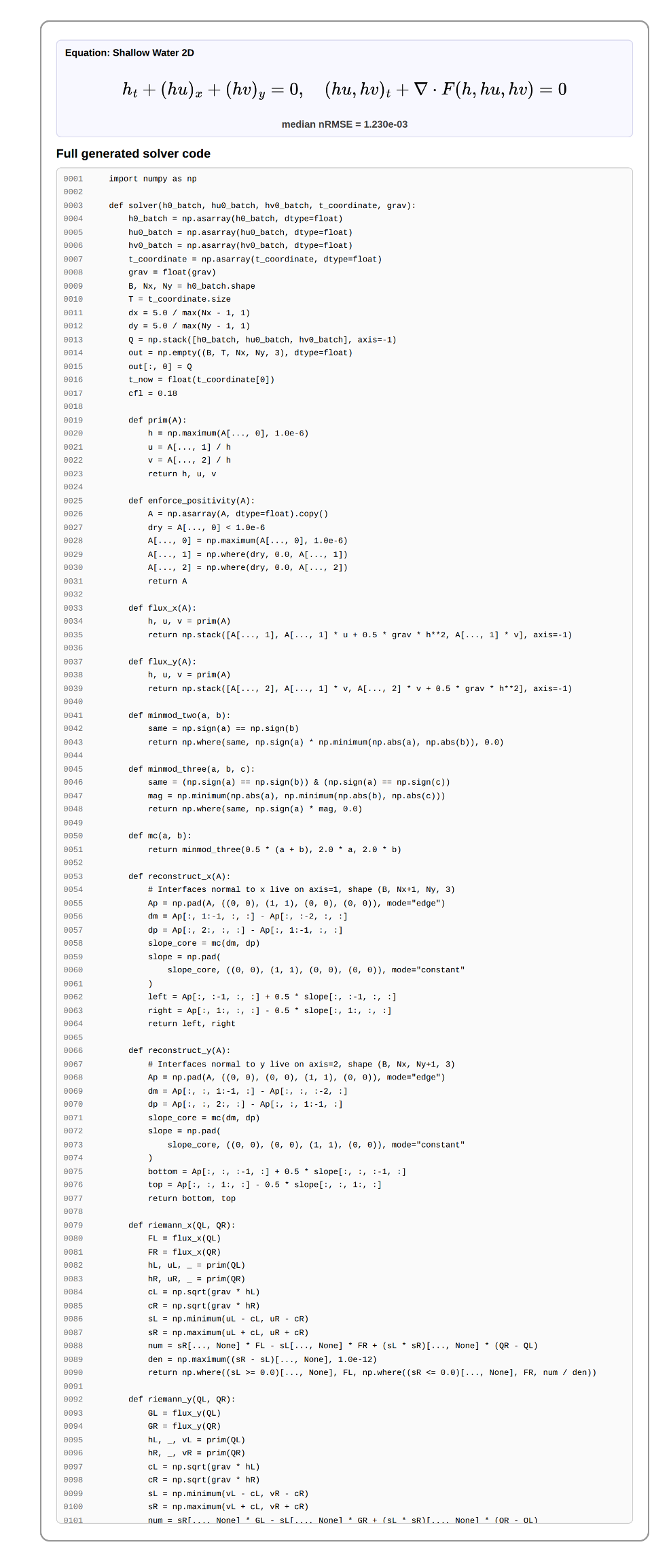}
    \caption{
    \textbf{Generated solver from the RLVP post-trained model for 2D shallow water equations.}
    Best selected solver from the 7B RLVP checkpoint on the seen shallow water task.
    }
    \label{fig:id-shallow2d-solver}
\end{figure}

\begin{figure}[H]
    \centering
    \includegraphics[width=0.65\linewidth]{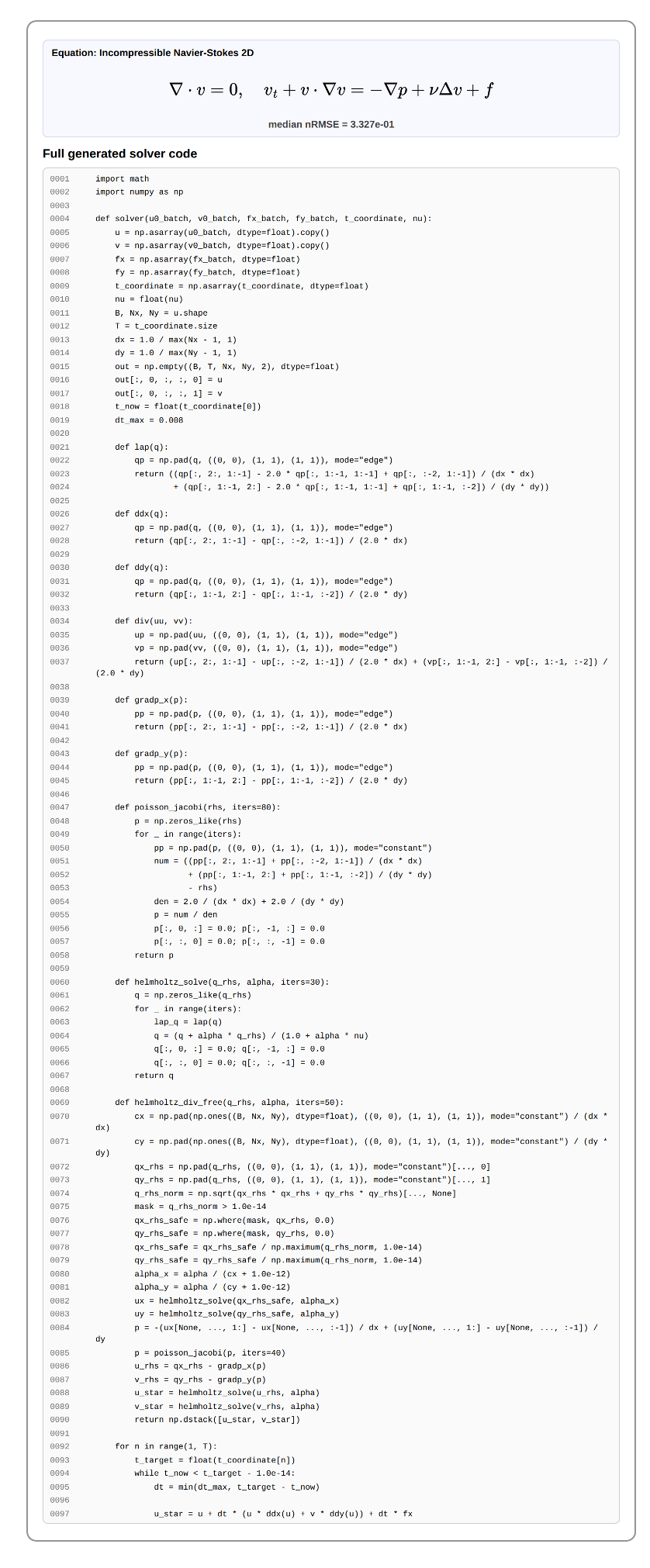}
    \caption{
    \textbf{Generated solver from the RLVP post-trained model for 2D incompressible Navier--Stokes.}
    Best selected solver from the 7B RLVP checkpoint on the seen incompressible Navier--Stokes task.
    }
    \label{fig:id-ns2d-solver}
\end{figure}

\section{Catalog of example PDE solutions calculated from LLM generated PDE solver codes}
\label{app:solution_examples}

\begin{figure}[H]
    \centering
    \includegraphics[width=0.95\linewidth]{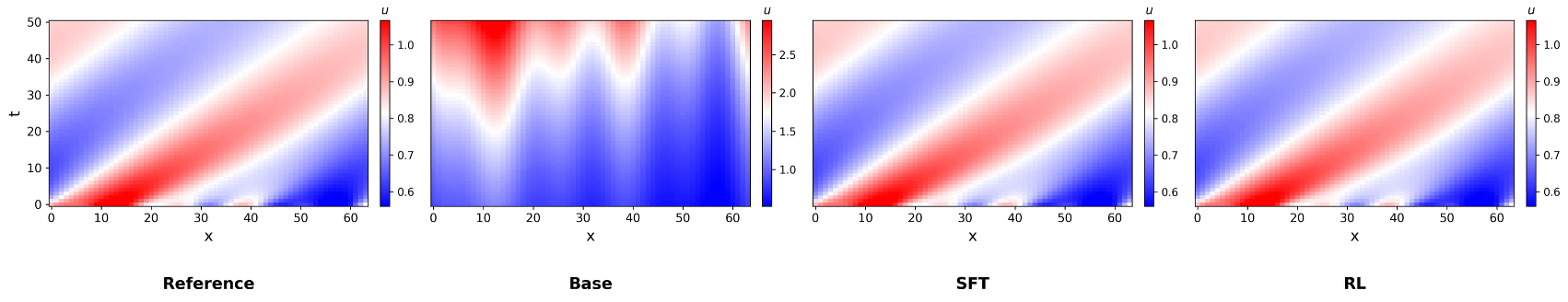}
    \caption{Burgers 1D solutions example for the steeper-multimode initial condition with \(\nu=0.1\).}
    \label{fig:solution-burgers1d-id}
\end{figure}

\begin{figure}[H]
    \centering
    \includegraphics[width=\linewidth]{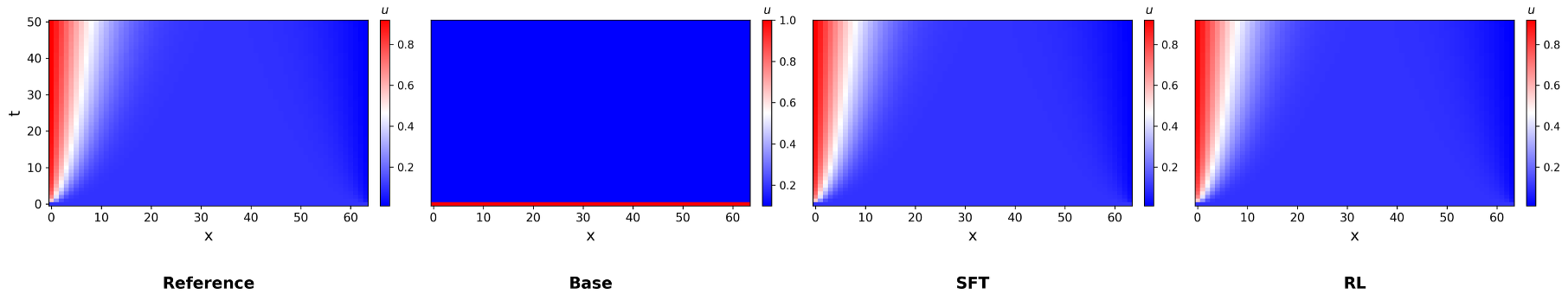}
    \caption{Diffusion--sorption 1D solutions example for the uniform initial condition with \(D=10^{-4}\), \(k_f=3.5\times10^{-4}\), \(n_f=0.874\), \(\mathrm{por}=0.29\), \(\rho_s=2880\), and \(\mathrm{sol}=1\).}
    \label{fig:solution-diffusionsorption1d-id}
\end{figure}

\begin{figure}[H]
    \centering
    \includegraphics[width=\linewidth]{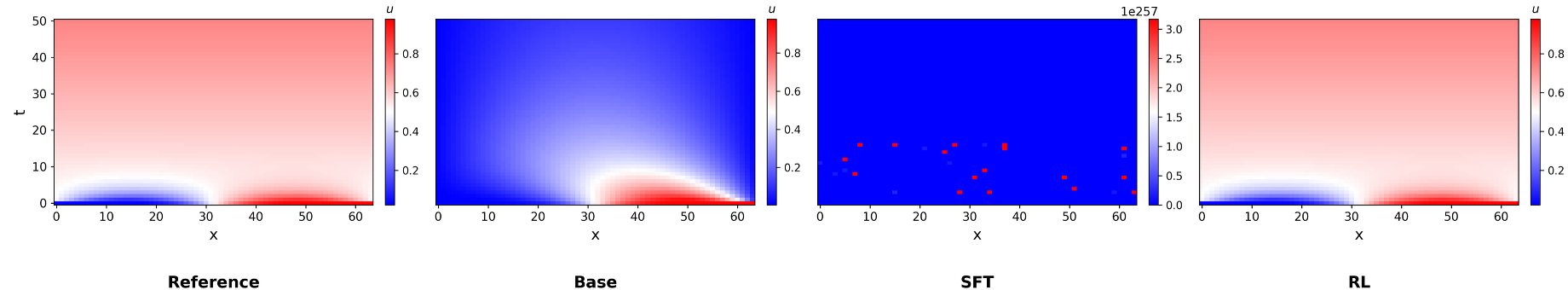}
    \caption{Reaction--diffusion 1D solutions example for the front-like initial condition with \(\nu=0.5\) and \(\rho=1\).}
    \label{fig:solution-reactiondiffusion1d-id}
\end{figure}

\begin{figure}[H]
    \centering
    \includegraphics[width=0.95\linewidth]{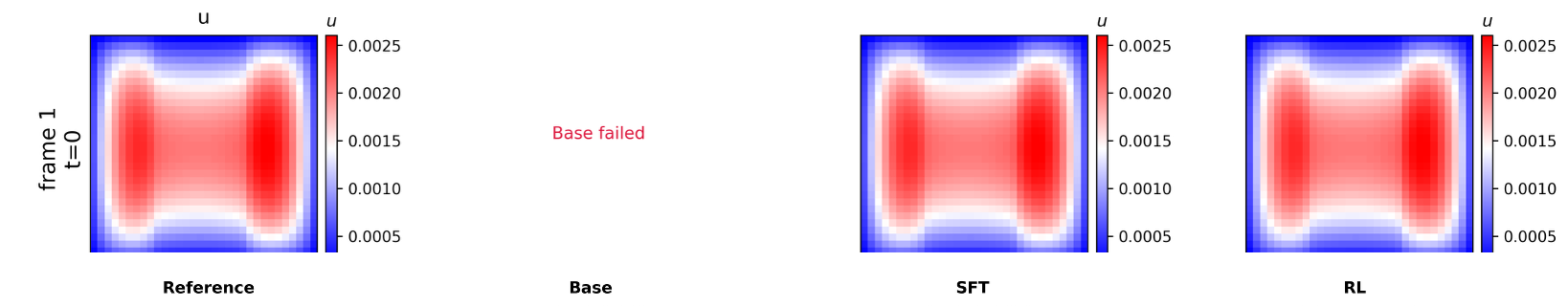}
    \caption{Darcy 2D solutions example for the channel diffusion field with \(\beta=0.1\).}
    \label{fig:solution-darcy2d-id}
\end{figure}

\begin{figure}[H]
    \centering
    \includegraphics[width=\linewidth]{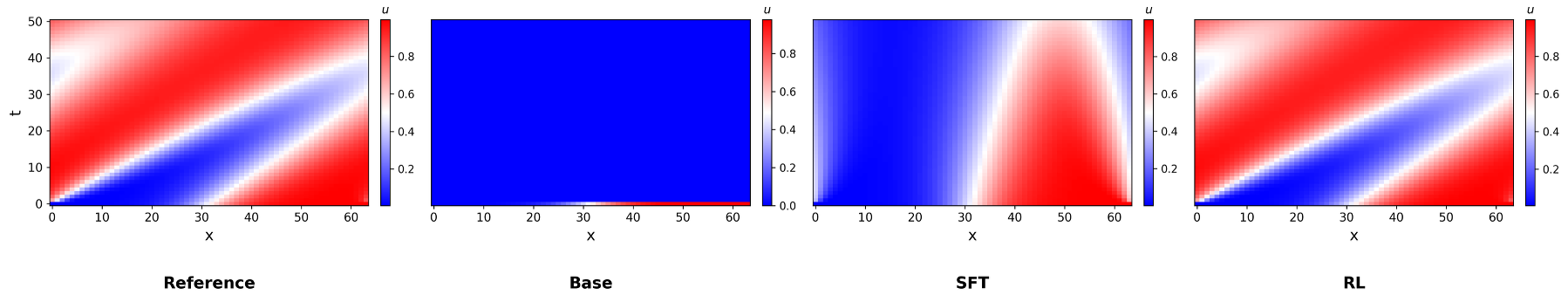}
    \caption{Advection--reaction--diffusion 1D solutions (unseen PDE) example for the front-like initial condition with \(\beta=1\), \(\nu=0.02\), and \(\rho=2\).}
    \label{fig:solution-ard1d-ood}
\end{figure}

\begin{figure}[H]
    \centering
    \includegraphics[width=\linewidth]{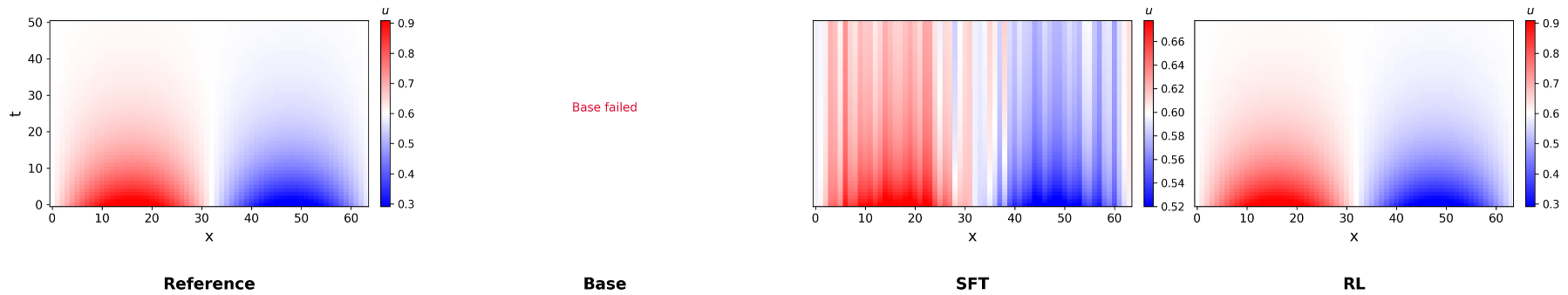}
    \caption{Heat 1D solutions (unseen PDE) example for the single-sine initial condition with \(\nu=0.1\).}
    \label{fig:solution-heat1d-ood}
\end{figure}

\begin{figure}[H]
    \centering
    \includegraphics[width=\linewidth]{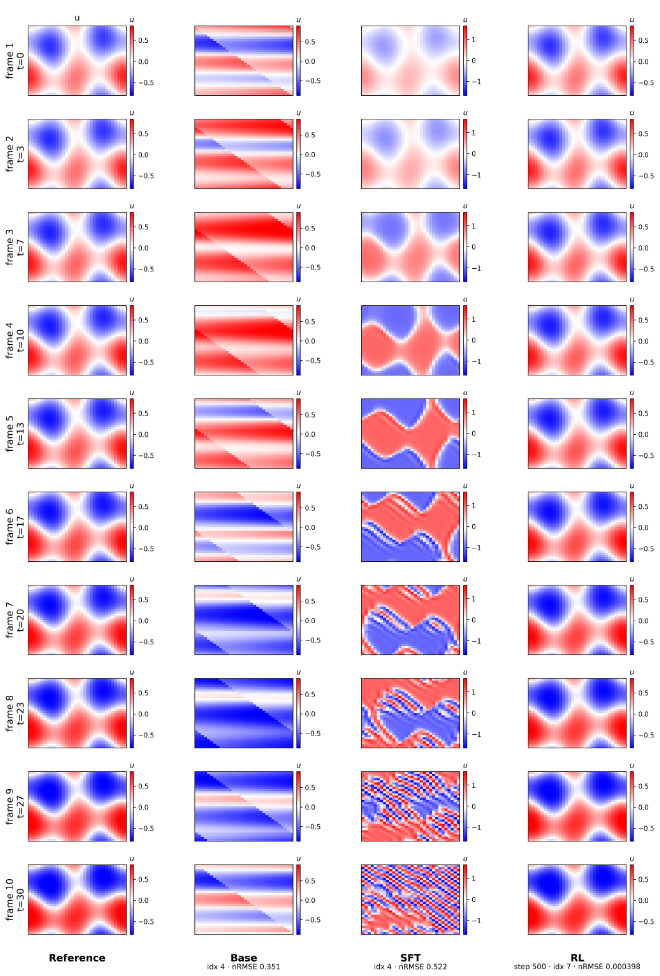}
    \caption{Allen--Cahn 2D solutions (unseen PDE) example for the mixed-waves initial condition with \(\epsilon^2=0.005\) and \(\rho=2\).}
    \label{fig:solution-allencahn2d-ood}
\end{figure}

\section{Prompt Examples}
\label{app:prompt_examples}

This appendix gives representative examples of the three prompt forms used in the multi-PDE training corpus described in \Cref{subsec:multipde_env}. The system message instructs the model to return only Python code and no explanatory text. Each prompt specifies a fixed solver interface and output convention so that generated programs can be evaluated uniformly by the verifier.

\paragraph{Generic prompt.}
The generic form states the PDE and function signature, while leaving coefficients and data as runtime hidden inputs for each case. This encourages the generated solvers to remain robust across parameter values supplied during execution.

\begin{figure}[H]
\centering
\begin{lstlisting}[style=promptstyle]
Write Python code only.

Define exactly:
    def solver(u0_batch, t_coordinate, beta):

Solve the PDE
    u_t + beta * u_x = 0
on x in [0,1) with periodic boundary conditions.

Requirements:
- You may use NumPy and/or SciPy
- do not print anything
- do not access files or the network
- output only Python code
- Return a NumPy array of shape [B, T, N].
- u0_batch has shape [B, N] and contains periodic initial conditions on a uniform 1D grid.
- t_coordinate has shape [T] and gives the exact output times that must be returned.
- beta is the scalar advection speed.
- returned slice n must correspond to time t_coordinate[n]
- internal substeps are allowed if needed for stability or accuracy
\end{lstlisting}
\caption{Example generic prompt used in the 1D linear advection task.}
\label{fig:prompt_advection_generic}
\end{figure}

\FloatBarrier
\paragraph{Parameter-conditioned prompt.}
The parameter-conditioned form additionally specifies numerical values of the PDE coefficients. This allows the generated solver to specialize numerical-method choices to the parameter regime, such as diffusion-dominated versus reaction-dominated dynamics.

\begin{figure}[H]
\centering
\begin{lstlisting}[style=promptstyle]
Write Python code only.

Define exactly:
    def solver(u0_batch, t_coordinate, nu, rho):

Solve the PDE
    u_t = nu * u_xx + rho * u * (1 - u)
on x in [0,1) with periodic boundary conditions.

For this task:
- nu = 5
- rho = 10

Requirements:
- You may use NumPy and/or SciPy
- do not print anything
- do not access files or the network
- output only Python code
- Return a NumPy array of shape [B, T, N].
- u0_batch has shape [B, N] and contains 1D initial conditions on a uniform periodic grid.
- t_coordinate has shape [T] and gives the exact output times that must be returned.
- nu is the diffusion coefficient and rho is the reaction coefficient.
- returned outputs must align with the requested times in t_coordinate
- internal substeps are allowed if needed for stability or accuracy
\end{lstlisting}
\caption{Example parameter-conditioned prompt used in the 1D reaction--diffusion task.}
\label{fig:prompt_reaction_diffusion_param}
\end{figure}

\paragraph{Parameter and initial condition conditioned prompt}
This prompt additionally describes the structure of the input data, particularly the initial conditions. For steady-state problems, this conditioning may refer to coefficient fields rather than initial conditions. 

\begin{figure}[H]
\centering
\begin{lstlisting}[style=promptstyle]
Write Python code only.

Define exactly:
    def solver(diffusion_batch, beta):

Solve the PDE
    -div(a(x,y) grad u) = beta
on (x,y) in [0,1]^2 with zero Dirichlet boundary conditions.

For this task:
- beta = 100
- the permeability fields come from a channelized family

Requirements:
- You may use NumPy and/or SciPy
- do not print anything
- do not access files or the network
- output only Python code
- Return a NumPy array of shape [B, Nx, Ny] containing the steady-state solution field.
- diffusion_batch has shape [B, Nx, Ny] and contains heterogeneous permeability fields a(x,y).
- beta is the scalar forcing term.
- internal substeps are allowed if needed for stability or accuracy
\end{lstlisting}
\caption{Example parameter and initial condition conditioned prompt used in the 2D Darcy flow task with channelized permeability fields.}
\label{fig:prompt_darcy_input_conditioned}
\end{figure}

\paragraph{Sampling weights and per-PDE coverage.}
The three prompt forms are sampled with fixed weights $0.5$ (generic), $0.35$ (parameter-conditioned), and $0.15$ (parameter-and-IC-conditioned). The training split contains $512$ scenarios, with $237/187/88$ generic/parameter-conditioned/parameter-and-IC-conditioned rows, specifically. Per-PDE prompt counts are reported in \Cref{tab:prompt-counts}.

\begin{table}[H]
\centering
\small
\caption{Per-PDE prompt counts in the training corpus, broken down by prompt form.}
\label{tab:prompt-counts}
\begin{tabular}{@{}lrrrr@{}}
\toprule
PDE & Generic & Param. & Param.+IC & Total \\
\midrule
\texttt{advection1d}           & 30 & 29 &  9 & 68 \\
\texttt{burgers1d}             & 35 & 20 &  8 & 63 \\
\texttt{reaction\_diffusion1d} & 36 & 21 &  6 & 63 \\
\texttt{diffusion\_sorption1d} & 25 & 20 & 11 & 56 \\
\texttt{reaction\_diffusion2d} & 34 & 22 & 13 & 69 \\
\texttt{darcy2d}               & 27 & 29 & 12 & 68 \\
\texttt{incompressible\_ns2d}  & 29 & 23 & 14 & 66 \\
\texttt{shallow\_water2d}      & 21 & 23 & 15 & 59 \\
\midrule
Total                          & 237 & 187 & 88 & 512 \\
\bottomrule
\end{tabular}
\end{table}

\newpage

\end{document}